\documentclass{article} 
\usepackage{iclr2027_conference,times}

\usepackage{amsmath,amsfonts,bm}

\def\eqref#1{equation~\ref{#1}}

\def\1{\bm{1}}

\DeclareMathAlphabet{\mathsfit}{\encodingdefault}{\sfdefault}{m}{sl}
\SetMathAlphabet{\mathsfit}{bold}{\encodingdefault}{\sfdefault}{bx}{n}

\definecolor{mycitecolor}{RGB}{1, 101, 202}
\usepackage[colorlinks=true,
    citecolor=mycitecolor,
    linkcolor=blue,
    urlcolor=black]
    {hyperref}
\usepackage{url}
\usepackage{tabularx}
\usepackage{array}
\usepackage[normalem]{ulem}
\usepackage{algorithm}
\usepackage{algpseudocode}
\usepackage{multirow}
\usepackage{booktabs}
\usepackage{graphicx}
\usepackage{makecell}
\usepackage[table]{xcolor}
\usepackage{arydshln}
\usepackage{amssymb}
\usepackage{subcaption}
\usepackage{wrapfig}
\usepackage{float}

\newcommand{\method}{T$^2$Mem}
\title{Test-Time Memory Construction}

\definecolor{modelrow}{RGB}{238,244,255}

\author{%
\textbf{Zijun Wang}$^{1}$\thanks{indicates equal contribution.}\quad
\textbf{Yifan Ji}$^{1}$\footnotemark[1]\quad
\textbf{Haidong Xin}$^{1}$\quad
\textbf{Zhenghao Liu}$^{1}$\thanks{indicates corresponding author.}\quad
\textbf{Yukun Yan}$^{2}$\quad
\textbf{Zulong Chen}$^{3}$\\
\textbf{Yu Gu}$^{1}$\quad
\textbf{Ge Yu}$^{1}$\quad
\textbf{Maosong Sun}$^{2}$\\[0.5em]
$^{1}$School of Computer Science and Engineering, Northeastern University, China\\
$^{2}$Department of Computer Science and Technology, Tsinghua University, China\\
$^{3}$Alibaba Group, China
}

\begin{document}

\maketitle
\lhead{Arxiv Preprint}

\begin{abstract}
Large Language Models (LLMs) struggle to answer questions over long interaction histories, where relevant evidence may be scattered across distant sessions. Existing memory systems typically construct and maintain sophisticated memory representations offline, before user queries are observed, incurring substantial overhead and limiting their ability to adapt memory construction to query-specific information needs. In this paper, we propose \method{}, a \textbf{T}est-\textbf{T}ime \textbf{Mem}ory construction framework that defers memory construction to inference and progressively builds memory along the reasoning process. Specifically, \method{} extracts lightweight episode-level metadata offline to capture key events and temporal associations as retrieval cues, while retaining links to the original interactions to facilitate the reconstruction of dependencies across episodes. At test time, the LLM adaptively retrieves and organizes metadata and raw interactions, progressively constructing query-specific memory as reasoning unfolds. Experiments on LoCoMo and LongMemEval show the effectiveness of \method{} by consistently outperforming strong baselines across different backbone LLMs while substantially reducing the overhead of memory construction. Further analysis demonstrates that deferring memory construction to inference enables the memory construction process to dynamically adapt to the information needs of individual queries. All code is available at \url{https://github.com/OpenBMB/T2Mem}.
\end{abstract}
\section{Introduction}
Large Language Models (LLMs) have demonstrated strong emergent capabilities across a wide range of tasks~\citep{yao2023react,li2025search,xin2026lisrec,liu2026knowledge}. However, as user interactions grow longer, LLMs still struggle to reason over extended interaction histories due to limited context windows and the difficulty of effectively locating relevant information within long contexts~\citep{hu2025memory,liu2024lost,hsieh2024ruler}. A straightforward solution is Retrieval-Augmented Generation (RAG), which segments historical interactions into chunks and retrieves relevant ones at inference time~\citep{lewis2020rag,edge2024graphrag,li2025search,li2026seek}. However, chunk-level retrieval often treats historical interactions as isolated units, potentially disrupting temporal context and cross-turn dependencies among related interactions. This fragmentation makes it difficult for LLMs to fully exploit the interconnected information embedded in long interaction histories~\citep{xin2026metamem,zhang2025survey}.

To address this limitation, recent memory systems~\citep{chhikara2025mem0,kang2025memoryos,fang2025lightmem} organize long interaction histories into manageable memory units. These methods introduce dedicated mechanisms to filter, summarize, and consolidate historical interactions into persistent memory representations, enabling efficient access in future interactions. However, constructing and maintaining such sophisticated memory representations can incur substantial overhead. In particular, offline memory construction must anticipate diverse future queries and preserve sufficiently comprehensive information in each memory unit, resulting in considerable preprocessing and maintenance costs that can hinder the scalability and practical deployment of memory systems~\citep{peng2026memshot}. Recent efforts have therefore explored deferring memory construction to test time, reducing the reliance on costly preconstructed memory representations~\citep{yu2026lazymem,yin2026defermem}. Nevertheless, these approaches typically construct memory from an initially retrieved candidate set, making it difficult to acquire additional evidence that is not retrieved initially but becomes necessary as the memory construction process unfolds.

Recent advances in test-time scaling have demonstrated the potential of leveraging additional inference-time computation to elicit stronger reasoning capabilities from LLMs. By allocating more computation at inference time, LLMs can iteratively explore, verify, and revise intermediate solutions~\citep{snell2024scaling,openai2026openaio1card,guo2025deepseek}. This raises a natural question: \textit{Can memory construction similarly benefit from inference-time reasoning, allowing an LLM to dynamically determine what historical information to activate and organize as it reasons?} We answer this question by treating memory not as a fixed representation constructed offline, but as an evolving state that is progressively constructed during inference. 

Based on this perspective, we propose \method{}, a \textbf{T}est-\textbf{T}ime \textbf{Mem}ory construction framework that dynamically constructs memory during inference. \method{} draws inspiration from associative recall, where relevant memories can be progressively activated as retrieval cues emerge~\citep{teyler2007hippocampal,horner2015holistic}. Instead of constructing sophisticated memory representations offline, \method{} first extracts lightweight episode-level metadata that captures salient events and temporal associations, while retaining backlinks to the corresponding raw interactions. The metadata provides a compact structure for associating related episodes and locating their underlying evidence. During inference, the LLM progressively reasons over the query and uses this evolving reasoning process to navigate the metadata and selectively activate corresponding raw interactions. In this way, memory construction and reasoning are interleaved: each newly activated piece of evidence can inform subsequent reasoning and, in turn, guide the construction of a more relevant memory state. The final memory is therefore not predetermined before inference, but emerges progressively from the interaction between reasoning and memory activation.

In our experiments, we evaluate \method{} on LoCoMo and LongMemEval with two backbone LLMs. Across both benchmarks, \method{} consistently outperforms strong memory and retrieval baselines under comparable total token budgets, demonstrating the effectiveness of dynamically constructing memory at test time. Further analysis reveals that deferring memory construction to inference allows \method{} to adapt the construction process to the specific information needs of each query. This advantage is particularly evident for queries that require deeper memory construction, such as multi-hop and temporal questions, where relevant evidence is distributed across multiple episodes and cannot be fully identified through a single initial retrieval step. Moreover, by leveraging episode-level metadata, \method{} enables the LLM to establish dependencies across episodes and trace these dependencies back to relevant source interactions during reasoning, facilitating more effective memory construction without the substantial overhead of maintaining sophisticated  memory units.
\section{Related Work}
Large Language Models (LLMs) struggle to reason over long-horizon interaction histories due to limited context windows~\citep{liu2024lost,hsieh2024ruler}. Retrieval-Augmented Generation (RAG) methods address this limitation by segmenting interaction histories into chunks and retrieving query-relevant ones as context for answering user queries~\citep {lewis2020rag,edge2024graphrag}. However, such chunk-based retrieval often breaks the contextual and temporal dependencies inherent in interaction histories~\citep{xin2026metamem}. Recent memory systems therefore introduce dedicated memory management mechanisms to organize, retain, and update historical interactions. For example, MemGPT~\citep{packer2023memgpt} and MemoryOS~\citep{kang2025memoryos} are inspired by operating systems and maintain memory states by continuously incorporating relevant information into existing memory units. MemoryBank~\citep{zhong2024memorybank} draws inspiration from human memory and incorporates the Ebbinghaus forgetting curve to regulate the retention and updating of memory units. Despite their effectiveness, these approaches primarily focus on memory maintenance, while the structure and representation of memory units, which can directly affect how effectively memories are retained, updated, and accessed, remain less explored.

To construct more effective memory representations, existing methods typically segment, compress, and consolidate interaction histories using different strategies. Specifically, HiMem~\citep{zhang2026himem} segments interaction histories along event boundaries to form coherent episodic memories, while LightMem~\citep{fang2025lightmem} compresses and consolidates interactions by topic to preserve salient information. Subsequent approaches further organize these compact memories into higher-level semantic~\citep{hu2026xmemory} or temporal~\citep{li2026timem} structures, allowing evidence distributed across interactions to be accessed at different levels of abstraction. To further enhance the traceability of memory representations, AnchorMem~\citep{shen2026anchormem} links memory units to their original interactions, allowing fine-grained source information to be retrieved when needed. However, these approaches often incur substantial overhead in constructing and maintaining sophisticated memory representations, which can hinder the practical deployment and scalability of memory systems~\citep{peng2026memshot}.

To alleviate the reliance of memory effectiveness on predefined memory representations, recent work has shifted toward query-time memory construction~\citep{yu2026lazymem,yin2026defermem}. These methods typically retrieve a broad set of query-relevant facts and reorganize them into memory. While this paradigm enables more comprehensive memory construction, its retrieval process remains largely rigid and one-shot, focusing on retrieving more relevant information rather than actively identifying what information is actually needed to construct an effective memory. Such a passive retrieval process may therefore fail to acquire missing evidence that only becomes apparent from the partially constructed memory~\citep{yao2023react,li2025search}. To enable LLMs to actively acquire information according to their evolving needs, \method{} treats memory construction as a dynamic reasoning process at test time. Inspired by the test-time scaling of LLMs~\citep{openai2026openaio1card,guo2025deepseek,snell2024scaling}, \method{} allows the evolving memory state to actively determine what evidence should be retrieved and integrated next. This closed-loop process enables the model to seek information genuinely needed for memory construction, better distinguish between potentially conflicting or ambiguous facts, and progressively refine memory as reasoning unfolds.
\section{Methodology}
\label{sec:method}
In this section, we introduce \method{}, a \emph{test-time memory construction} framework for long-term conversational question answering. Unlike conventional approaches that construct a query-agnostic memory before inference, \method{} dynamically constructs memory at test time by progressively uncovering and organizing relevant information from the raw interaction history. We first revisit existing memory systems and show how \method{} decouples memory construction from the offline preprocessing stage, enabling memory to be constructed dynamically during inference (Sec.~\ref{sec:preliminaries}). We then present the architecture of \method{}, which leverages the reasoning capabilities of LLMs to progressively incorporate relevant information to construct tailored memory (Sec.~\ref{sec:memory_construction}).

\subsection{Disentangling Memory Orchestration from Offline to Test Time}
\label{sec:preliminaries}
Given a long-term interaction history $\mathcal{S}=\{s_1,s_2,\ldots,s_n\}$, a memory system organizes information scattered across interaction sessions into manageable memory units $\mathcal{M}=\{m_1,m_2,\ldots,m_N\}$. These memory units focus on preserving factual knowledge, user preferences, and cross-session dependencies by resolving conflicts in the interactions and incorporating potentially useful information. Existing memory systems typically construct such memory representations offline, before future queries are presented to the LLM. In contrast, \method{} defers memory orchestration to test time, using the given query to incorporate acquired information into the reasoning trajectory, thereby organizing relevant knowledge dynamically during inference.

\textbf{Offline Memory Construction.}
Existing memory systems first construct a query-agnostic memory collection $\mathcal{M}$ from the raw interaction history $\mathcal{S}$ by extracting, summarizing, and organizing potentially useful information:
\begin{equation}
\label{offline_memory}
\mathcal{M}=\pi_{\mathrm{mem}}(s_1,s_2,\ldots,s_n),
\end{equation}
where $\pi_{\mathrm{mem}}$ denotes the memory construction policy, which may repeatedly invoke LLMs to extract, update, and delete information as new interactions arrive, while also maintaining sophisticated memory representations such as hierarchical or graph-based structures. Moreover, since such construction is query-agnostic, the memory system must organize information without knowing which parts will be relevant to future queries, often incurring substantial construction and maintenance overhead~\citep{peng2026memshot}. Given a user query $q$, the system retrieves a subset of the preconstructed memory units and uses them to generate the answer:
\begin{equation}
\label{offline_answer}
y=\operatorname{LLM}(q,\operatorname{Retriever}(q,\mathcal{M})),
\end{equation}
where $\operatorname{Retriever}(\cdot)$ denotes the memory retrieval procedure. By extracting and organizing information in advance, these methods maintain persistent memory representations that can be directly retrieved at inference time~\citep{kang2025memoryos,hu2026evermemos}.

\textbf{Test-Time Memory Evolving.}
Rather than constructing sophisticated memory representations offline, \method{} mimics human associative recall by progressively activating relevant episodes as reasoning unfolds, allowing the LLM to orchestrate memory at test time~\citep{horner2015holistic}. To represent episodic experiences~\citep{fang2025lightmem}, \method{} first extracts episode-level metadata that captures the dependencies among key events and temporal associations within each episode, using concise episode descriptions as retrieval cues. Given a query, the LLM then progressively uncovers query-relevant information by following these cues and locating the corresponding raw dialogue through the extracted metadata. 

Specifically, \method{} employs an LLM $\pi_{\mathrm{meta}}$ to extract lightweight, episode-level metadata $\mathcal{I}$ from the dialog history $\mathcal{S}=\{s_1,s_2,\ldots,s_n\}$:
\begin{equation}
\mathcal{I}=\pi_{\mathrm{meta}}(s_1,s_2,\ldots,s_n),
\end{equation}
where $\mathcal{I}=\{i_1,i_2,\ldots,i_n\}$ contains extracted metadata entries for dialog sessions. Each $i_k$ organizes the key events, facts, entities, and temporal cues from session $s_k$ into a semi-structured representation, while maintaining a source link $\mathcal{L}(i_k)=s_k$ for traceability~\citep{sun2026trajwiki,shen2026anchormem}. These metadata entries serve as compact retrieval cues for associating related episodes at test time (Eq.~\ref{eq:testtime_answer}), allowing \method{} to identify cross-episode dependencies through the metadata and trace the retrieved cues back to their corresponding source sessions via $\mathcal{L}(i_k)$.

Subsequently, given a query $q$, \method{} dynamically constructs memory by accessing the metadata $\mathcal{I}$ and the original interaction history $\mathcal{S}$ through a set of information-access tools $\mathcal{T}$:
\begin{equation}
\label{eq:testtime_answer}
y=\operatorname{LLM}\bigl(q,\mathcal{I},\mathcal{S},\mathcal{T}\bigr).
\end{equation}
Rather than relying on rigid, one-shot retrieval, \method{} invokes the tools in $\mathcal{T}$ during inference to progressively acquire relevant information in response to its evolving reasoning needs. The retrieved information is incorporated into the ongoing reasoning process, enabling the model to iteratively construct memory as reasoning unfolds. Thus, \method{} decouples memory construction from costly offline memory orchestration, instead deferring memory construction to test time, where it can be dynamically tailored to the query and the evolving reasoning process.

\begin{figure}[t]
\centering
\small
\includegraphics[width=\linewidth]{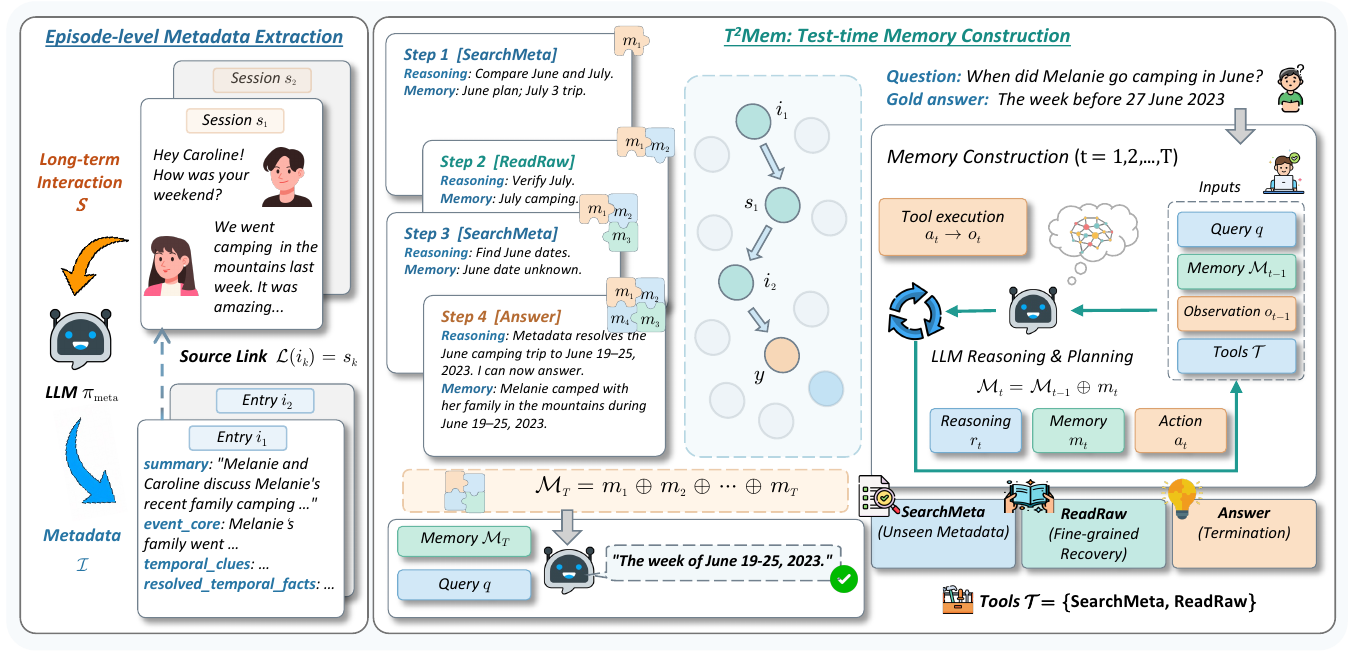}
\caption{Overall Architecture of Our Proposed \method{} Framework.}
\label{fig:model}

\end{figure}

\subsection{Dynamic Memory Construction During LLM Reasoning}
\label{sec:memory_construction}
As illustrated in Figure~\ref{fig:model}, we detail how \method{} dynamically constructs query-specific memory at test time (Eq.~\ref{eq:testtime_answer}) by progressively acquiring, distilling, and organizing relevant information during reasoning. At each step, the LLM reasons over the accumulated memory and the current observation, distills query-relevant information into the evolving memory state, and determines what additional information is needed. Based on this information need, it either terminates the process or selects an appropriate tool from $\mathcal{T}$ to access episode-level metadata or fine-grained raw information, thereby supporting subsequent memory construction.

We first formulate this process as an iterative reasoning trajectory $\tau$ consisting of $T$ memory construction steps:
\begin{equation}
\tau=\bigl((r_1,m_1,a_1),(r_2,m_2,a_2),\ldots,(r_T,m_T,a_T)\bigr),
\end{equation}
where $r_t$, $m_t$, and $a_t$ denote the reasoning trace, newly constructed memory cache, and selected action at step $t$, respectively. The accumulated memory caches $\mathcal{M}_t$ generated throughout the trajectory are progressively concatenated to form the final query-specific memory $\mathcal{M}_T$:
\begin{equation}
\label{final_memory}
\mathcal{M}_T=m_1\oplus m_2\oplus\cdots\oplus m_T,
\end{equation}
where $\oplus$ denotes the concatenation operation. The LLM then generates the final answer conditioned on the user query $q$ and the resulting memory $\mathcal{M}_T$:
\begin{equation}
\label{TTMC_answer}
y=\operatorname{LLM}(q,\mathcal{M}_T).
\end{equation}
\textbf{Dynamic Memory Construction.} We next describe how each step produces $(r_t,m_t,a_t)$ and updates the memory state. At step $t$, the LLM takes the query $q$, the accumulated memory $\mathcal{M}_{t-1}$, the latest observation $o_{t-1}$, and the available tool set $\mathcal{T}$ as input, and jointly produces a reasoning trace $r_t$, a newly constructed memory cache $m_t$, and the next action $a_t$:
\begin{equation}
\label{eq:memory_construction}
(r_t,m_t,a_t)=\operatorname{LLM}(q,\mathcal{M}_{t-1},o_{t-1},\mathcal{T}),
\end{equation}
where the memory is initialized as $\mathcal{M}_0=\varnothing$. The initial observation is obtained by retrieving the top-$k$ metadata entries in $\mathcal{I}$ most relevant to $q$:
\begin{equation}
o_0=\operatorname{Top-}k(q,\mathcal{I}),
\end{equation}
where $\operatorname{Top-}k$ selects the $k$ metadata entries in $\mathcal{I}$ that are most relevant to the query $q$. The reasoning trace $r_t$ guides the LLM in distilling query-relevant evidence from $o_{t-1}$ into the newly constructed memory cache $m_t$. The memory is then updated by appending $m_t$ to the accumulated memory:
\begin{equation}
\mathcal{M}_t=\mathcal{M}_{t-1}\oplus m_t,
\end{equation}
thereby completing the memory update at step $t$. The process terminates when the current memory state satisfies the information need of the given query or when the maximum number of construction steps $T_{\max}$ is reached. Let $T\leq T_{\max}$ denote the actual number of executed steps. Finally, the accumulated memory $\mathcal{M}_T$ is used to help the LLM generate the final answer according to Eq.~\ref{TTMC_answer}.

\textbf{Adaptive Searching During LLM Reasoning.}
After constructing $m_t$, the LLM executes the selected action $a_t$, either terminating the process with \texttt{Answer} or invoking an information-access tool to obtain the next observation $o_t$. Specifically, we define the tool set $\mathcal{T}$ as:
\begin{equation} 
\mathcal{T}=\left\{\texttt{SearchMeta}(\cdot),\texttt{ReadRaw}(\cdot)\right\}, 
\end{equation}
where $\texttt{SearchMeta}(\cdot)$ retrieves additional episode-level metadata, while $\texttt{ReadRaw}(\cdot)$ follows the source link $\mathcal{L}(i_k)=s_k$ associated with the observed metadata entry $i_k$ to recover the corresponding raw session. To avoid repeatedly reading the same metadata entry, we maintain a query-specific set of previously observed entries, $\mathcal{I}^{\mathrm{seen}}\subseteq\mathcal{I}$, initialized with the metadata entries contained in $o_0$.

When the current evidence is insufficient and additional events or temporal cues are required, the LLM selects $\texttt{SearchMeta}(\cdot)$ to acquire previously unseen metadata. The tool excludes entries already contained in $\mathcal{I}^{\mathrm{seen}}$ and retrieves the highest-ranked remaining entries:
\begin{equation}
\texttt{SearchMeta}(q)=\operatorname{Top-}k\left(q,\mathcal{I}\setminus\mathcal{I}^{\mathrm{seen}}\right).
\end{equation}
After each call to $\texttt{SearchMeta}(\cdot)$, the returned entries are added to $\mathcal{I}^{\mathrm{seen}}$, preventing redundant reading.
When an observed metadata entry $i_k$ points to a relevant session but does not provide sufficient detail, the LLM can instead select $\texttt{ReadRaw}(k)$ to access the corresponding raw session:
\begin{equation}
\label{eq:read_raw}
\texttt{ReadRaw}(k)=\mathcal{L}(i_k)=s_k,
\end{equation}
where $i_k\in\mathcal{I}^{\mathrm{seen}}$. This operation enables the LLM to recover fine-grained evidence from the corresponding raw interaction when the lightweight metadata is insufficient, thereby allowing the memory construction process to adaptively incorporate fine-grained evidence into the evolving memory.

\section{Experimental Methodology}
In this section, we describe the experimental setup for evaluating \method{}. We first introduce the datasets, baselines, and evaluation metrics, and then present the implementation details.

\begin{table}[t]
\centering
\small \caption{Overall Performance of \method{} and Baseline Methods on the LoCoMo Dataset. The \textbf{best} and \underline{second-best} results are highlighted.}\label{tab:overall_locomo} \resizebox{\linewidth}{!}{
\begin{tabular}{lrccccc}
\toprule
\textbf{Method} & \textbf{\# Token ($\times10^3$)} & \textbf{Multi-Hop} & \textbf{Temporal} & \textbf{Open-Domain} & \textbf{Single-Hop} & \textbf{Avg.}\\
\midrule

\rowcolor{gray!20}
\multicolumn{7}{c}{\itshape\textbf{Qwen3.5-27B}}\\
\midrule

Naive RAG~\citeyearpar{lewis2020rag} & 1,012 & 32.98 & 60.12 & 38.54 & 64.68 & 56.30\\
Search-o1~\citeyearpar{li2025search} & 2,380 & 49.29 & 61.68 & \underline{53.12} & 60.64 & 58.31\\
MemoryOS~\citeyearpar{kang2025memoryos} & 6,460 & 51.77 & 41.12 & 42.71 & 72.77 & 60.45\\
LightMem~\citeyearpar{fang2025lightmem} & 6,178 & 58.51 & 70.72 & 40.62 & 74.55 & 68.70\\
EverMemOS~\citeyearpar{hu2026evermemos} & 12,955 & 51.77 & 66.36 & \textbf{58.33} & 69.20 & 64.74\\
HiMem~\citeyearpar{zhang2026himem} & 11,056 & \underline{59.03} & 68.37 & 49.81 & 71.17 & 65.34\\
xMemory~\citeyearpar{hu2026xmemory} & 8,229 & 57.09 & \underline{71.96} & 47.92 & 72.65 & 68.12\\
AnchorMem~\citeyearpar{shen2026anchormem} & 10,940 & 47.16 & 70.72 & 42.71 & 75.83 & 67.45\\

LazyMem~\citeyearpar{yu2026lazymem} & 11,253 & 51.77 & 62.50 & 36.96 & \textbf{82.28} & \underline{69.84}\\

\method{} (Ours) & 8,765 & \textbf{59.22} & \textbf{75.70} & 50.00 & \underline{76.81} & \textbf{71.69}\\
\midrule
\rowcolor{gray!20}
\multicolumn{7}{c}{\itshape\textbf{Gemma4-31B-Instruct}}\\
\midrule

Naive RAG~\citeyearpar{lewis2020rag} & 1,025 & 32.62 & 59.50 & 31.25 & 63.50 & 55.00\\
Search-o1~\citeyearpar{li2025search} & 1,891 & 39.36 & 57.63 & 44.79 & 58.03 & 53.70\\
MemoryOS~\citeyearpar{kang2025memoryos} & 7,357 & 56.03 & 64.17 & \underline{50.00} & 75.51 & 65.84\\
LightMem~\citeyearpar{fang2025lightmem} & 6,326 & \textbf{62.06} & \textbf{77.26} & 37.50 & 72.06 & 69.16\\
EverMemOS~\citeyearpar{hu2026evermemos} & 11,063 & 52.48 & 65.42 & \underline{50.00} & 66.47 & 62.66\\
HiMem~\citeyearpar{zhang2026himem} & 10,289 & 47.87 & 66.36 & 28.12 & 63.81 & 59.19\\
xMemory~\citeyearpar{hu2026xmemory} & 8,190 & 55.32 & 71.34 & 45.83 & 71.94 & 67.14\\
AnchorMem~\citeyearpar{shen2026anchormem} & 8,949 & 48.94 & \underline{76.01} & 40.62 & 74.52 & 68.03\\

LazyMem~\citeyearpar{yu2026lazymem} & 10,624 & 52.48 & 62.50 & 36.96 & \textbf{83.23} & \underline{70.49}\\

\method{} (Ours) & 7,274 & \underline{56.74} & \textbf{77.26} & \textbf{52.08} & \underline{76.10} & \textbf{71.30}\\

\bottomrule
\end{tabular}
}
\end{table}

\textbf{Datasets.}
We evaluate \method{} on two widely adopted benchmarks for long-term memory, LoCoMo~\citep{maharana2024locomo} and LongMemEval~\citep{wu2025longmemeval}. These benchmarks assess the ability of memory systems to retain, retrieve, and utilize information from long-term interaction histories. Further details are provided in Appendix~\ref{app:dataset}.

\textbf{Baselines.}
We compare \method{} with several representative baselines spanning three categories: retrieval-based methods, memory systems that construct structured memory representations offline, and recent test-time approaches that defer memory construction to inference.

For retrieval-based methods, we evaluate Naive RAG~\citep{lewis2020rag} and Search-o1~\citep{li2025search}, which directly retrieve relevant dialogue segments to support LLM-based answering. We then compare \method{} with representative memory systems. MemoryOS~\citep{kang2025memoryos} draws inspiration from operating systems and continuously incorporates relevant information into existing memory units. LightMem~\citep{fang2025lightmem} and HiMem~\citep{zhang2026himem} construct compact memory units through topic-based consolidation and event-level organization, respectively. EverMemOS~\citep{hu2026evermemos} and xMemory~\citep{hu2026xmemory} further organize memory into higher-level semantic structures to facilitate subsequent retrieval. AnchorMem~\citep{shen2026anchormem} preserves links between memory units and their source interactions, enabling the recovery of fine-grained details when needed. Finally, we adopt LazyMem~\citep{yu2026lazymem} as a test-time baseline, which follows a retrieve-then-construct paradigm to build memory from a broad set of retrieved candidates during inference.

\textbf{Evaluation Metrics.}
Following prior work~\citep{fang2025lightmem,xin2026metamem}, we use accuracy as the primary evaluation metric and employ an LLM-based judge to determine whether each response is consistent with the ground-truth answer~\citep{zheng2023judging}. For a fair comparison, all methods are evaluated using Qwen3.5-Plus~\citep{qwen3.5} as the judge model. The evaluation prompts are provided in Appendix~\ref{app:prompt}.

\textbf{Implementation Details.}
In our experiments, we use Qwen3.5-27B~\citep{qwen3.5} and Gemma4-31B-Instruct~\citep{team2026gemma} as the backbone models for all compared methods, and BGE-M3~\citep{bge-m3} as the embedding model for metadata retrieval. For \method{}, we set $k=5$ for the initial retrieval from $\mathcal{I}$ to obtain $o_0$. In each subsequent \texttt{SearchMeta} call, we set $k=1$ to retrieve the highest-ranked entry from $\mathcal{I}\setminus\mathcal{I}^{\mathrm{seen}}$. We set the maximum number of memory construction steps to $T_{\max}=5$ in the main experiments. The prompt templates used by \method{} are provided in Appendix~\ref{app:prompt}, while additional implementation details are reported in Appendix~\ref{app:experiment}.

\section{Experimental Results}
In this section, we first report the overall performance of \method{} and conduct ablation studies to show the effectiveness of different components. Then, we evaluate the quality of the memory constructed by \method{} and other representative methods. Subsequently, we compare token consumption between offline preprocessing and test-time inference, and investigate the test-time scaling of \method{} under different test-time token budgets. Finally, case studies are provided in Appendix~\ref{app:case}.

\begin{table}[t]
\centering
\small \caption{Overall Performance of \method{} and Baseline Methods on the LongMemEval Dataset.} 
\label{tab:overall_lme} 
\resizebox{\linewidth}{!}{
\begin{tabular}{lccccccc}
\toprule
\multirow{2}{*}{\textbf{Method}} & \textbf{Single} & \textbf{Single}& \textbf{Single} & \textbf{Multi} & \textbf{Knowledge} & \textbf{Temporal} & \multirow{2}{*}{\textbf{Avg.}} \\
& \textbf{User} & \textbf{Assistant} & \textbf{Preference} &\textbf{Session} & \textbf{Update} & \textbf{Reasoning}  &  \\

\midrule
\rowcolor{gray!20}
\multicolumn{8}{c}{\itshape\textbf{Qwen3.5-27B}}\\
\midrule

Naive RAG~\citeyearpar{lewis2020rag} & 92.86 & 87.50 & 36.67 & 69.92 & 79.49 & 58.65 & 71.60\\
Search-o1~\citeyearpar{li2025search} & 88.57 & 89.29 & 40.00 & 57.14 & 75.64 & 59.40 & 67.60\\

MemoryOS~\citeyearpar{kang2025memoryos} & 90.00 & 82.14 & 36.67 & 60.15 & 69.23 & 48.87 & 63.80\\
LightMem~\citeyearpar{fang2025lightmem} & 91.43 & 25.00 & \underline{76.67} & \textbf{78.20} & 87.18 & \textbf{85.71} & \underline{77.40}\\
EverMemOS~\citeyearpar{hu2026evermemos} & 72.68 & 80.36 & 33.33 & 55.64 & 70.51 & 52.63 & 61.00\\

HiMem~\citeyearpar{zhang2026himem} & 75.71 & 50.00 & 33.33 & 54.14 & 70.51 & 56.39 & 58.60\\
xMemory~\citeyearpar{hu2026xmemory} & \underline{94.29} & 64.29 & 46.67 & 60.90 & 83.33 & 51.65 & 68.80\\
AnchorMem~\citeyearpar{shen2026anchormem} & \textbf{95.71} & \textbf{98.21} & 40.87 & 57.89 & 82.05 & 44.36 & 64.40\\

LazyMem~\citeyearpar{yu2026lazymem} & 92.86 & \underline{91.07} & 63.33 & 52.63 & \textbf{94.87} & 63.16 & 72.60\\
\method{} (Ours) & 92.86 & \textbf{98.21} & \textbf{83.33} & \underline{70.68} & \underline{92.31} & \underline{84.96} & \textbf{84.80}\\
\midrule
\rowcolor{gray!20}
\multicolumn{8}{c}{\itshape\textbf{Gemma4-31B-Instruct}}\\
\midrule

Naive RAG~\citeyearpar{lewis2020rag} & \textbf{97.14} & 85.71 & 56.67 & \underline{73.68} & \underline{93.59} & 63.16 & \underline{77.60}\\
Search-o1~\citeyearpar{li2025search} & 94.29 & 94.64 & 33.33 & 62.41 & 73.08 & 62.41 & 70.40\\

MemoryOS~\citeyearpar{kang2025memoryos} & 92.86 & 82.14 & 30.00 & 69.17 & 79.49 & 60.15 & 70.80\\
LightMem~\citeyearpar{fang2025lightmem} & 92.86 & 10.71 & \textbf{90.00} & \textbf{79.70} & 87.18 & \textbf{87.22} & \underline{77.60}\\
EverMemOS~\citeyearpar{hu2026evermemos} & 88.57 & 78.57 & 60.00 & 70.68 & 88.46 & 62.41 & 74.00\\

HiMem~\citeyearpar{zhang2026himem} & 82.86 & 55.36 & 40.00 & 60.15 & 76.92 & 62.41 & 64.80\\
xMemory~\citeyearpar{hu2026xmemory} & 94.29 & 66.07 & 50.00 & 60.90 & 83.33 & 61.65 & 69.20\\
AnchorMem~\citeyearpar{shen2026anchormem} & \underline{95.71} & \underline{98.21} & 36.67 & 69.17 & \textbf{94.87} & 63.91 & 76.80\\

LazyMem~\citeyearpar{yu2026lazymem} & \underline{95.71} & 96.43 & 73.33 & 55.64 & 91.03 & 69.92 & 76.20\\
\method{} (Ours) & 94.29 & \textbf{100.00} & \underline{83.33} & 69.92 & 92.31 & \underline{85.71} & \textbf{85.20}\\
\bottomrule
\end{tabular}
}
\end{table}
\subsection{Overall Performance}
\label{sec:overall}
In this subsection, we evaluate the overall performance of \method{} against a range of representative baselines. The results on LoCoMo are presented in Table~\ref{tab:overall_locomo}, with additional evaluation on LongMemEval reported in Table~\ref{tab:overall_lme}. Overall, \method{} outperforms all baselines across both benchmarks, demonstrating the effectiveness of our approach to dynamically constructing memory at test time. Notably, \method{} delivers consistent gains across different backbone models, indicating its robustness and generalizability across model architectures.

Compared with retrieval-based methods such as Naive RAG and Search-o1, \method{} achieves higher average performance on both benchmarks. On LoCoMo, the gains are particularly evident on Multi-Hop and Temporal questions, suggesting that dynamically constructing memory according to evolving information needs is more effective than relying on a fixed set of retrieved chunks or iterative retrieval alone. \method{} also exhibits more consistent performance across task categories than offline memory construction methods. These methods compress and organize the interaction history into persistent memory representations to enable efficient access in future interactions, but the preconstructed memory may not align equally well with different information needs. For example, on LongMemEval (Table~\ref{tab:overall_lme}), LightMem shows strong performance on several task categories but substantially lower accuracy on Single-Assistant questions. This suggests that such offline memory construction can lead to substantial performance variation across different task categories.

Furthermore, \method{} also outperforms LazyMem, which constructs memory from a one-pass retrieval set of query-relevant facts, making subsequent construction dependent on the quality and coverage of the initial retrieval. In contrast, \method{} unfolds memory construction along the reasoning process, progressively retrieving the evidence needed to construct and refine the memory. This enables \method{} to adaptively acquire missing evidence as reasoning unfolds, resulting in more comprehensive memory construction. Moreover, by supplying the necessary evidence at the appropriate stages of reasoning, \method{} avoids redundant reasoning and achieves stronger performance.

\begin{table}[t]
\centering
\small \caption{Ablation Study on the LoCoMo Dataset.} \label{tab:ablation_locomo} \resizebox{\linewidth}{!}{
\begin{tabular}{lrccccc}
\toprule

\textbf{Method} & \textbf{\# Token ($\times10^3$)} & \textbf{Multi-Hop} & \textbf{Temporal} & \textbf{Open-Domain} & \textbf{Single-Hop} & \textbf{Avg.}\\
\midrule

\rowcolor{gray!20}
\multicolumn{7}{c}{\textit{\textbf{Qwen3.5-27B}}}\\
\midrule

\method{} (Summary) & 6,809 & 57.80 & 71.96 & \underline{51.04} & 73.01 & 68.64\\
\method{} (Graph) & 7,943 & 52.13 & 59.19 & 45.83 & 69.92 & 62.92\\
\method{} & 8,765 & \underline{59.22} & \underline{75.70} & 50.00 & \textbf{76.81} & \textbf{71.69}\\
\hdashline
w/o Metadata & 11,792 & \textbf{59.92} & 68.84 & 47.91 & \underline{75.86} & \underline{69.74}\\ 
w/o Rawdata & 7,415 & 56.03 & \textbf{77.88} & \textbf{56.52} & 68.85 & 67.60\\ 
w/o Rollout Construction & 5,177 & 55.67 & 71.34 & 42.71 & 69.08 & 65.45\\
\midrule

\rowcolor{gray!20}
\multicolumn{7}{c}{\textit{\textbf{Gemma4-31B-Instruct}}}\\
\midrule

\method{} (Summary) & 5,245 & \underline{54.26} & \textbf{79.13} & 48.96 & 75.74 & \underline{70.84}\\
\method{} (Graph) & 7,133 & 52.48 & 67.91 & 43.75 & 70.04 & 64.74\\
\method{} & 7,274 & \textbf{56.74} & \underline{77.26} & \textbf{52.08} & \underline{76.10} & \textbf{71.30}\\
\hdashline
w/o Metadata & 11,831 & 52.48 & 72.90 & \underline{50.00} & \textbf{78.48} & \underline{70.84}\\
w/o Rawdata & 6,668 & \underline{54.26} & \textbf{79.13} & \underline{50.00} & 66.94 & 66.10\\
w/o Rollout Construction & 3,641 & 52.84 & 75.39 & 42.71 & 63.85 & 62.92\\

\bottomrule
\end{tabular}
}
\end{table}

\subsection{Ablation Studies}
\label{sec:ablation}
This subsection conducts ablation studies on LoCoMo to examine the contributions of different components in \method{}. Additional results on LongMemEval are reported in Appendix~\ref{longmemeval_ablation}.

Specifically, we first examine the effectiveness of the lightweight episode-level metadata proposed in Sec.~\ref{sec:preliminaries} by comparing it with two alternative representations. \method{} (Summary) replaces the metadata for each session with a concise summary~\citep{tan2025prospect}, while \method{} (Graph) represents the relations among entities and events within each session using a graph structure~\citep{wu2026finding}. We further investigate the roles of metadata and raw interactions during memory construction. \method{} (w/o Metadata) directly performs dynamic memory construction over the raw interaction sessions without using metadata, whereas \method{} (w/o Rawdata) relies solely on the metadata. Finally, \method{} (w/o Rollout Construction) removes the rollout memory construction process and directly answers the user query using the initial retrieval results. As shown in Table~\ref{tab:ablation_locomo}, \method{} consistently outperforms both \method{} (Summary) and \method{} (Graph), demonstrating the effectiveness of its semi-structured metadata. This advantage can be attributed to the complementary structural and semantic cues, such as timestamps, event information, and topics, which help LLMs capture dependencies among events during reasoning and identify relevant sessions for memory construction. Furthermore, the performance degradation in \method{} (w/o Metadata) and \method{} (w/o Rawdata) highlights the complementary roles of these two knowledge sources. The metadata provides cues for locating relevant interactions, while the raw sessions preserve fine-grained events for recovering detailed evidence. Removing the rollout memory construction process reduces token consumption, but leads to a clear performance degradation in \method{} (w/o Rollout Construction). This suggests that progressively integrating retrieved knowledge according to evolving information needs enables more effective and tailored memory construction.

\begin{figure}[t]
\centering
\small
\begin{subfigure}[t]{0.32\linewidth}
\centering
\includegraphics[width=\linewidth]
{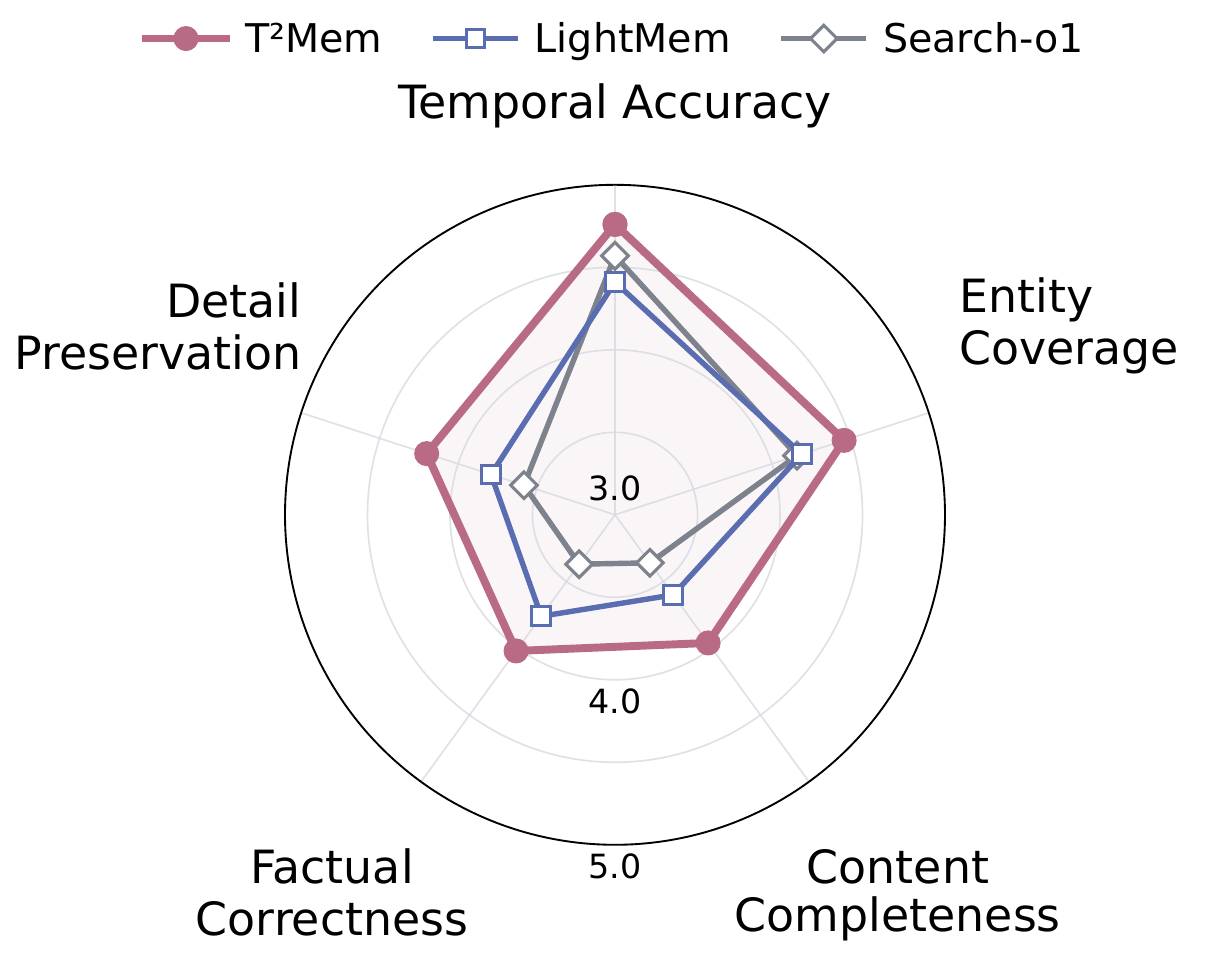}
\caption{Quality of the Constructed Memory.}
\label{fig:memory_quality_radar}
\end{subfigure}
\hfill
\begin{subfigure}[t]{0.32\linewidth}
\centering
\includegraphics[width=\linewidth]
{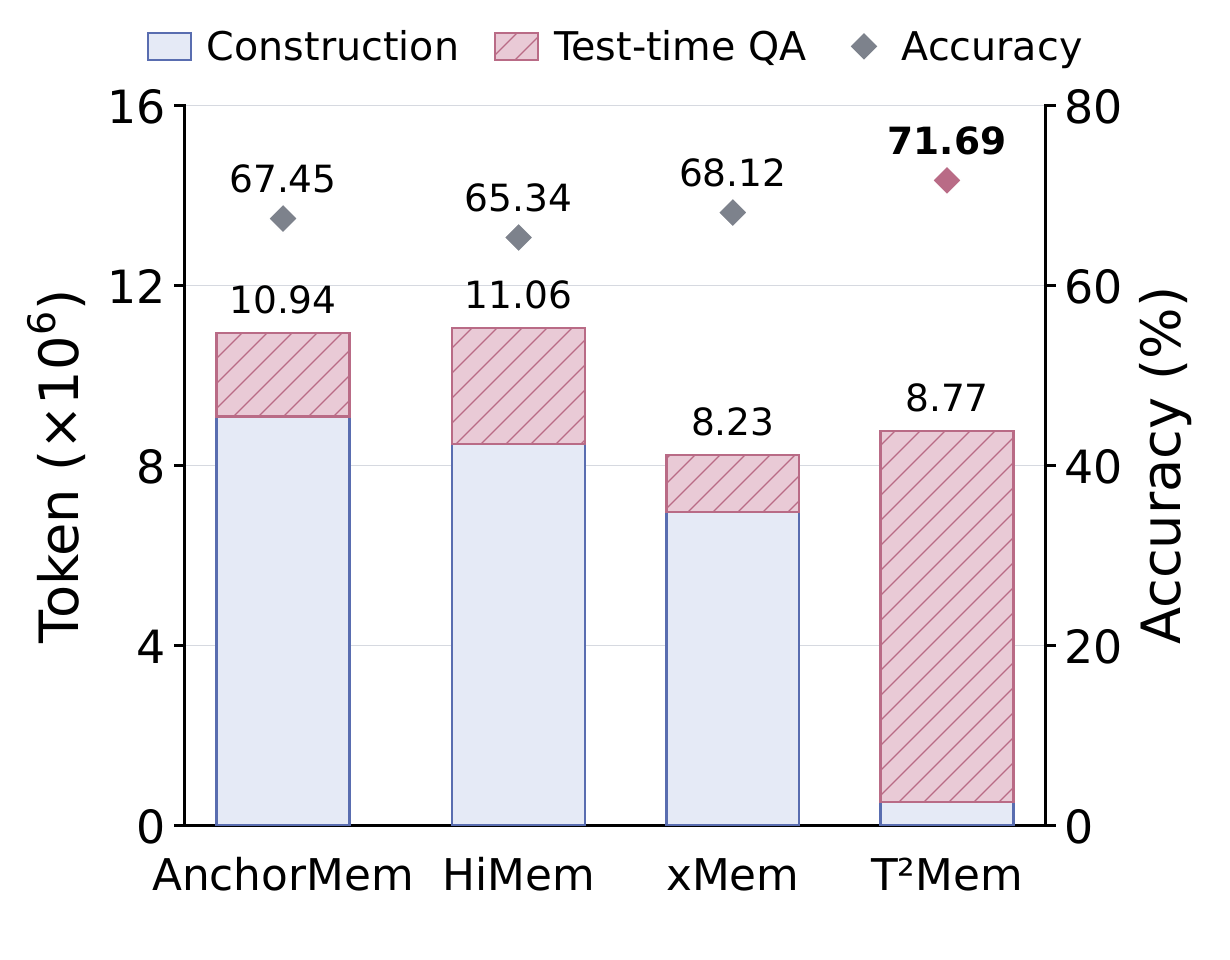}
\caption{Token Consumption for Memory Construction and Inference.}

\label{fig:token_consumption_comparison}
\end{subfigure}
\hfill
\begin{subfigure}[t]{0.32\linewidth}
\centering
\includegraphics[width=\linewidth]
{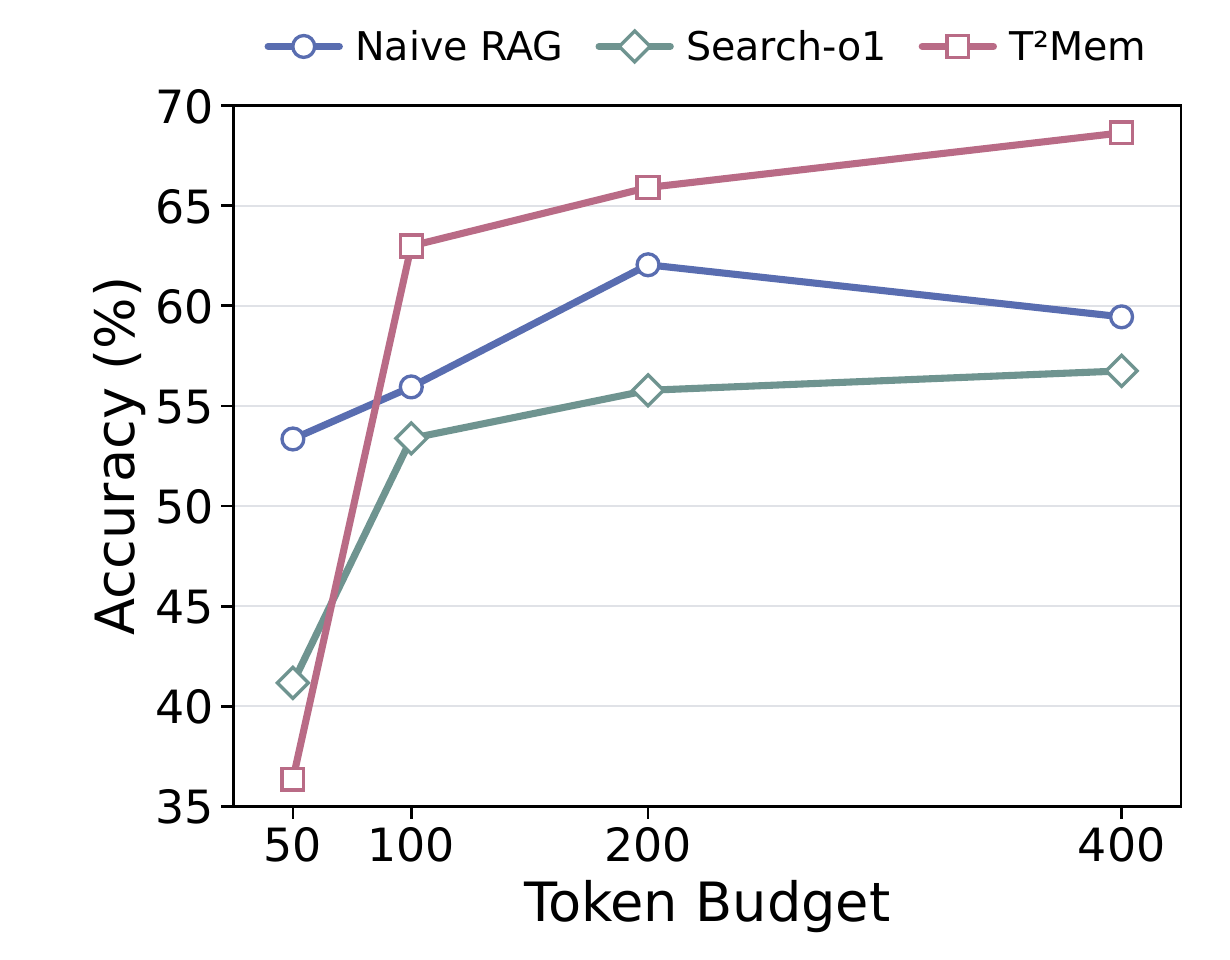}
\caption{QA Accuracy across Different Token Budgets.}
\label{fig:token_budget_accuracy}
\end{subfigure}

\caption{Analysis of \method{}. We examine (a) the quality of the constructed memory, (b) token consumption during offline memory construction and LLM inference, and (c) QA performance across different test-time token budgets.
\label{fig:memory_construction_characteristics}}

\end{figure}

\subsection{Effectiveness of Test-Time Computation for Memory Construction}

In this subsection, we analyze \method{} in terms of memory quality, token allocation, and test-time scaling, as illustrated in Figure~\ref{fig:memory_construction_characteristics}. All experiments are conducted using Qwen3.5-27B on the LoCoMo dataset. Additional experimental details are provided in Appendix~\ref{app:experiment}.

\textbf{Quality of Constructed Memory.}
As shown in Figure~\ref{fig:memory_quality_radar}, we evaluate the quality of the memory constructed by \method{}, LightMem, and Search-o1 along five dimensions, with the evaluation criteria detailed in Appendix~\ref{app:prompt}. Specifically, \textit{Temporal Accuracy} and \textit{Entity Coverage} assess the ability to capture temporal dependencies and entity associations across interactions. \textit{Content Completeness} and \textit{Factual Correctness} measure whether the constructed memory contains sufficient information to answer the query and remains consistent with the ground-truth answer, respectively. Finally, \textit{Detail Preservation} evaluates how well the memory retains specific quantities and other details. Across all five dimensions, \method{} achieves the highest scores, indicating that it constructs more complete and reliable memory. This advantage can be attributed to its ability to progressively capture episode dependencies, thereby helping retrieve more accurate evidence to facilitate the LLM in answering the query. In particular, the improvement in \textit{Detail Preservation} suggests that \method{} retains finer-grained evidence beyond the core information needed to answer the query. This benefit stems from its iterative construction process, which allows the model to read additional metadata or access linked source sessions when the accumulated memory is insufficient to support construction. The higher \textit{Factual Correctness} further demonstrates that \method{} can effectively incorporate supporting evidence, leading to better consistency with the ground-truth answers.

\textbf{Association between Token Consumption and QA Accuracy.}
As shown in Figure~\ref{fig:token_consumption_comparison}, \method{} achieves the highest QA accuracy while consuming a comparable number of tokens to the three offline memory construction systems. This efficiency stems from a different computation allocation strategy: rather than constructing comprehensive memories offline, \method{} extracts only lightweight metadata in advance and shifts the remaining computation to test time, allowing memory construction to adapt to the information needs of each query. To further investigate this property and its scaling behavior, we vary the token budget available to \method{} and compare it with Naive RAG and Search-o1. As shown in Figure~\ref{fig:token_budget_accuracy}, \method{} consistently improves with additional test-time computation, whereas Search-o1 exhibits a weaker scaling trend and Naive RAG shows no consistent benefit, with its performance even declining as the token budget increases. These results indicate that effective test-time scaling depends not merely on increasing the computation budget, but on effectively translating additional computation into useful evidence for memory evolution.
\section{Conclusion}
This paper presents \method{}, a test-time memory construction framework for long-term conversational question answering. \method{} extracts lightweight episode-level metadata offline as retrieval cues while retaining links to the original interactions, and adaptively retrieves and organizes metadata and raw interactions at test time to progressively construct query-specific memory. Experiments on LoCoMo and LongMemEval show that \method{} consistently outperforms strong baselines across two backbone LLMs while substantially reducing memory construction overhead. Further analysis demonstrates that deferring memory construction to inference enables the memory construction process to dynamically adapt to the information needs of individual queries.

\subsection*{AI use statement}
Generative AI tools were used only for proofreading, including the correction of typographical and grammatical errors. All methodological designs, experimental execution, interpretation of the results, manuscript content, references, and citations were prepared and verified by the authors. The authors take full responsibility for the final content of this work.


\bibliography{iclr2027_conference}
\bibliographystyle{iclr2027_conference}
\clearpage
\appendix
\section{Appendix}

\subsection{License}
All datasets used in this work are publicly available with licenses permitting academic research and non-commercial use. Specifically, LoCoMo is distributed under the Creative Commons Attribution-NonCommercial 4.0 International License (CC BY-NC 4.0), while LongMemEval is released under the MIT License.

\begin{table}[H]
\centering
\caption{Statistics of the LoCoMo and the LongMemEval Datasets.} \label{tab:dataset} \small
\begin{tabular}{lrlr}
\toprule
\multicolumn{2}{c}{\textbf{LoCoMo}} & \multicolumn{2}{c}{\textbf{LongMemEval}}\\
\cmidrule(lr){1-2}
\cmidrule(lr){3-4}
\textbf{Category} & \textbf{\# Questions} & \textbf{Category} & \textbf{\# Questions}\\
\midrule
Single-Hop & 841 & Single-Session User & 70\\
Multi-Hop & 282 & Single-Session Assistant & 56\\
Temporal & 321 & Single-Session Preference & 30\\
Open Domain & 96 & Multi-Session & 133\\
- & - & Knowledge-Update & 78\\
- & - & Temporal-Reasoning & 133\\
\midrule
Total & 1,540 & Total & 500\\
\bottomrule
\end{tabular}
\end{table}
\subsection{Additional Dataset Details}
\label{app:dataset}
Table~\ref{tab:dataset} summarizes the statistics of the two long-term memory benchmarks adopted in our experiments: LoCoMo~\citep{maharana2024locomo} and LongMemEval~\citep{wu2025longmemeval}. Following prior studies~\citep{fang2025lightmem,hu2026evermemos}, we exclude the \texttt{adversarial} category from LoCoMo for evaluation, and use the official \texttt{longmemeval\_s\_cleaned} split\footnote{\url{https://huggingface.co/datasets/xiaowu0162/longmemeval}} for LongMemEval.

\subsection{Additional Experimental Details}
\label{app:experiment}
In all experiments, we employ vLLM\footnote{\url{https://vllm.ai/}} to serve both the backbone and retrieval models. For generation, we set the temperature to 0.0, top-$p$ to 0.95, and top-$k$ to 20, with a maximum answer length of 1024 tokens. For HiMem~\citep{zhang2026himem}, we follow the original implementation and adopt bge-reranker-v2-m3~\citep{bge-m3} as the reranking model. For LazyMem~\citep{yu2026lazymem}, we replace the original memory-processing model, LazyMem-4B, with the corresponding backbone model, since LazyMem-4B had not been publicly released when our experiments were conducted. The reported LazyMem results therefore correspond to this backbone-substituted implementation.

For the token-budget analysis, we adopt prompt-based length control following prior work~\citep{han2025token,lee2025well}. Specifically, the model is instructed to generate at most $N$ reasoning tokens before producing the final answer. We override the default output limit by setting \texttt{max\_tokens} to $N+64$, where the additional 64 tokens provide a fixed allowance for the final answer and output formatting. This limit applies to the complete generated response and does not separately enforce the reasoning length. The final-answer instructions remain unchanged across all budget settings. Native thinking mode is disabled, and reasoning is generated explicitly in the response.

\begin{table}[t]
\centering
\small
\caption{Ablation Study on the LongMemEval Dataset.}
\label{tab:ablation_lme}
\resizebox{\linewidth}{!}{
\begin{tabular}{lccccccc}
\toprule
\multirow{2}{*}{\textbf{Method}}
& \textbf{Single}
& \textbf{Single}
& \textbf{Single}
& \textbf{Multi}
& \textbf{Knowledge}
& \textbf{Temporal}
& \multirow{2}{*}{\textbf{Avg.}} \\
& \textbf{User}
& \textbf{Assistant}
& \textbf{Preference}
& \textbf{Session}
& \textbf{Update}
& \textbf{Reasoning} \\
\midrule

\rowcolor{gray!20}
\multicolumn{8}{c}{\itshape\textbf{Qwen3.5-27B}} \\ 
\midrule

\method{} (Summary) & 85.71 & 94.64 & 60.00 & 60.90 & 82.05 & 73.68 & 74.80 \\
\method{} (Graph) & 84.29 & \textbf{100.00} & 70.00 & 59.40 & 82.05 & \underline{79.70} & 77.00 \\
\method{} & \underline{92.86} & \underline{98.21} & \textbf{83.33} & \underline{70.68} & \underline{92.31} & \textbf{84.96} & \textbf{84.80} \\
\hdashline
w/o Metadata & 90.00 & 94.64 & 53.33 & 62.41 & 89.74 & 75.19 & 77.00 \\
w/o Rawdata & 91.43 & \underline{98.21} & \underline{76.67} & \textbf{72.93} & 89.74 & \underline{79.70} & \underline{83.00} \\
w/o Rollout Construction & \textbf{94.29} & 94.64 & 73.33 & 69.17 & \textbf{93.59} & \underline{79.70} & 82.40 \\
\midrule

\rowcolor{gray!20}
\multicolumn{8}{c}{\itshape\textbf{Gemma4-31B-Instruct}} \\
\midrule

\method{} (Summary) & 91.43 & \textbf{100.00} & \underline{76.67} & 62.41 & 80.77 & 70.68 & 76.60 \\
\method{} (Graph) & 91.43 & \textbf{100.00} & \underline{76.67} & \underline{70.68} & \underline{91.03} & 81.95 & 83.40 \\
\method{} & \underline{94.29} & \textbf{100.00} & \textbf{83.33} & 69.92 & \textbf{92.31} & \textbf{85.71} & \textbf{85.20} \\
\hdashline
w/o Metadata & \underline{94.29} & \underline{94.64} & 56.67 & \textbf{73.68} & 89.74 & \underline{84.96} & \underline{84.00} \\
w/o Rawdata & \textbf{95.71} & \underline{94.64} & \underline{76.67} & 69.17 & \underline{91.03} & 83.46 & 83.40 \\
w/o Rollout Construction & \underline{94.29} & 91.07 & 73.33 & 66.92 & \underline{91.03} & 80.45 & 81.20 \\
\bottomrule
\end{tabular}
}
\end{table}
\subsection{Ablation Studies on the LongMemEval Dataset}
\label{longmemeval_ablation}
As shown in Table~\ref{tab:ablation_lme}, \method{} achieves the highest average accuracy on LongMemEval with both backbone models. Replacing episode-level metadata with summaries or graphs reduces average performance, supporting the effectiveness of the proposed representation. Removing either metadata or access to raw interactions also lowers average accuracy, highlighting their complementary roles in memory construction. Removing rollout construction leads to clear declines on Temporal Reasoning and Single Preference questions. These results further support progressively constructing memory according to the query's evolving information needs.

\subsection{Effectiveness of \method{} at Different Memory Construction Steps}
\label{sec:reconstruction_steps}
We conduct two analyses on LoCoMo to examine the effect of memory construction steps. We first vary the maximum construction budget $T_{\max}$ to evaluate its effect on answer accuracy. We then examine the actual number of steps $T$ and compare \method{} with Search-o1 across observed stopping steps. For \method{}, each construction step generates $(r_t,m_t,a_t)$ and appends the new memory cache $m_t$ to the accumulated memory.

\begin{figure*}[t]
\centering
\small

\begin{subfigure}[t]{0.24\textwidth}
    \centering
    \includegraphics[width=\linewidth]{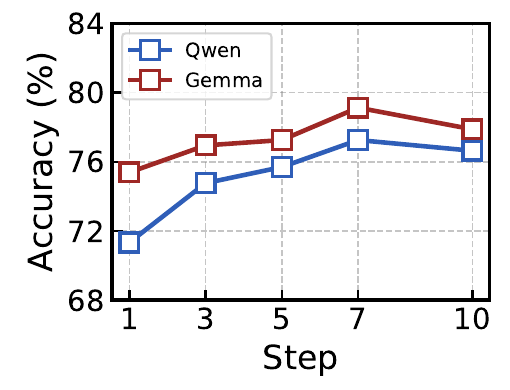}
    \caption{Temporal}
    \label{fig:locomo_step_temporal}
\end{subfigure}
\hfill
\begin{subfigure}[t]{0.24\textwidth}
    \centering
    \includegraphics[width=\linewidth]{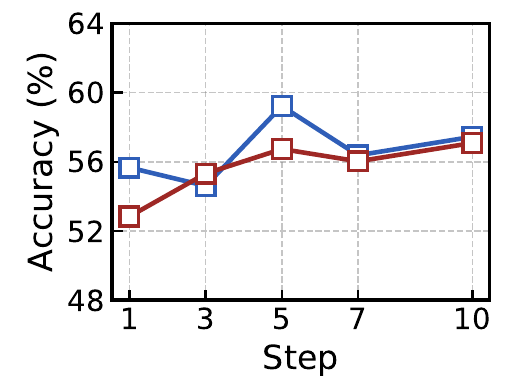}
    \caption{Multi-Hop}
    \label{fig:locomo_step_multihop}
\end{subfigure}
\hfill
\begin{subfigure}[t]{0.24\textwidth}
    \centering
    \includegraphics[width=\linewidth]{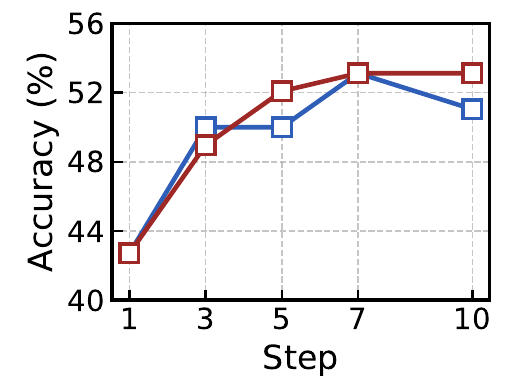}
    \caption{Open-Domain}
    \label{fig:locomo_step_opendomain}
\end{subfigure}
\hfill
\begin{subfigure}[t]{0.24\textwidth}
    \centering
    \includegraphics[width=\linewidth]{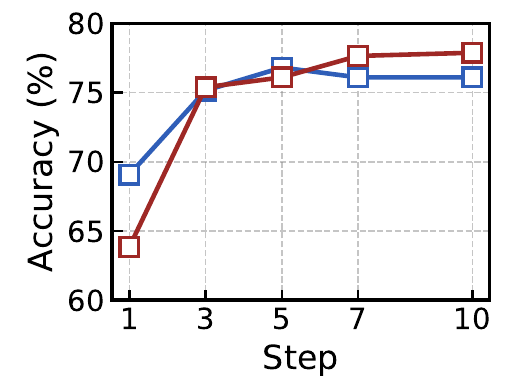}
    \caption{Single-Hop}
    \label{fig:locomo_step_singlehop}
\end{subfigure}

\caption{Performance of \method{} at Different Memory Construction Steps on the LoCoMo Dataset.}
\label{fig:locomo_step_analysis}
\end{figure*}
\textbf{Effect of Construction Steps on Answer Performance.}
As shown in Figure~\ref{fig:locomo_step_analysis}, the answer accuracy generally improves as the maximum number of construction steps increases, indicating that additional test-time computation benefits memory construction.  Specifically, Single-Hop questions show clear gains when the number of steps increases from one to three, followed by a clear plateau, as their required evidence can often be recovered within relatively shallow construction. Temporal and Open-Domain questions exhibit a different trend, with performance continuing to improve as the construction depth increases to seven steps. These questions typically require resolving temporal relations or recovering evidence distributed across interaction sessions, leaving more room for additional construction steps to contribute. Beyond this depth, the gains become less pronounced, reflecting diminishing returns from further extending the construction process.

\begin{figure*}[!t]
\centering
\small

\begin{subfigure}[t]{0.24\textwidth}
    \centering
    \includegraphics[width=\linewidth]{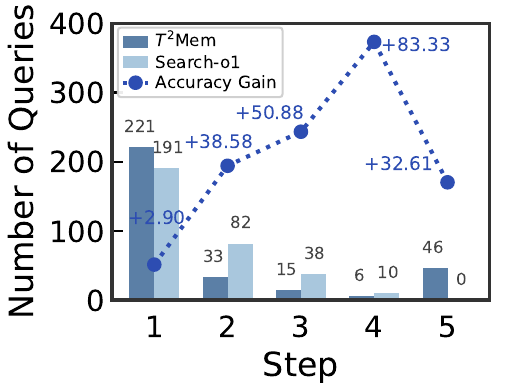}
    \caption{Temporal}
    \label{fig:test_time_reasoning_setting_1}
\end{subfigure}
\hfill
\begin{subfigure}[t]{0.24\textwidth}
    \centering
    \includegraphics[width=\linewidth]{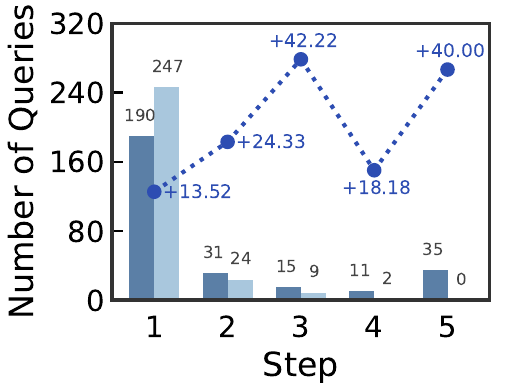}
    \caption{Multi-Hop}
    \label{fig:test_time_reasoning_setting_2}
\end{subfigure}
\hfill
\begin{subfigure}[t]{0.24\textwidth}
    \centering
    \includegraphics[width=\linewidth]{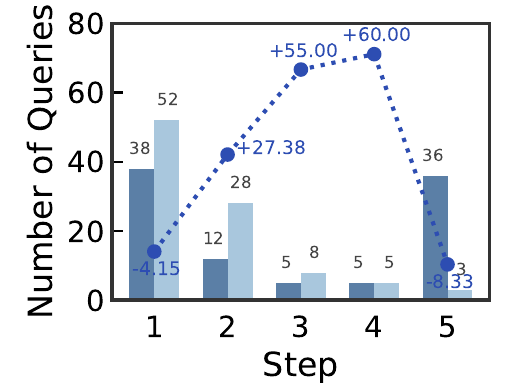}
    \caption{Open-Domain}
    \label{fig:test_time_reasoning_setting_3}
\end{subfigure}
\hfill
\begin{subfigure}[t]{0.24\textwidth}
    \centering
    \includegraphics[width=\linewidth]{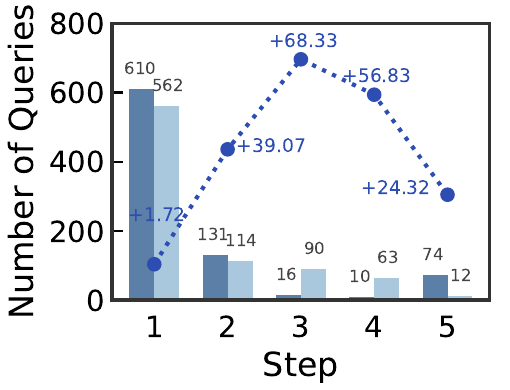}
    \caption{Single-Hop}
    \label{fig:test_time_reasoning_setting_4}
\end{subfigure}

\caption{Performance Gains of \method{} over Search-o1 across Different Memory Construction Steps on the LoCoMo Dataset.}
\label{fig:test_time_reasoning_effectiveness}
\end{figure*}
\textbf{Performance Gains of \method{} over Search-o1 at Different Construction Steps.}
Figure~\ref{fig:test_time_reasoning_effectiveness} further compares \method{} and Search-o1 according to the number of construction steps taken before answering. At the first step, the performance gap remains relatively small, particularly for Single-Hop questions, where the initially retrieved evidence is often sufficient to answer. However, as the process extends to two and three steps, \method{} achieves substantially larger gains across all types of questions. Although Search-o1 performs iterative retrieval, its subsequent searches are directly driven by the current reasoning state, without an explicit structure for organizing previously encountered events, entities, and temporal cues. In contrast, \method{} uses the lightweight episode-level metadata to provide structured cues for locating relevant sessions and selectively recovers fine-grained details from the linked source interactions as construction proceeds. This structured access allows successive steps to more effectively connect newly acquired evidence with information established earlier, which becomes increasingly beneficial when answering requires evidence scattered across multiple sessions.

\begin{figure}[t]
\centering
\includegraphics[width=\linewidth]{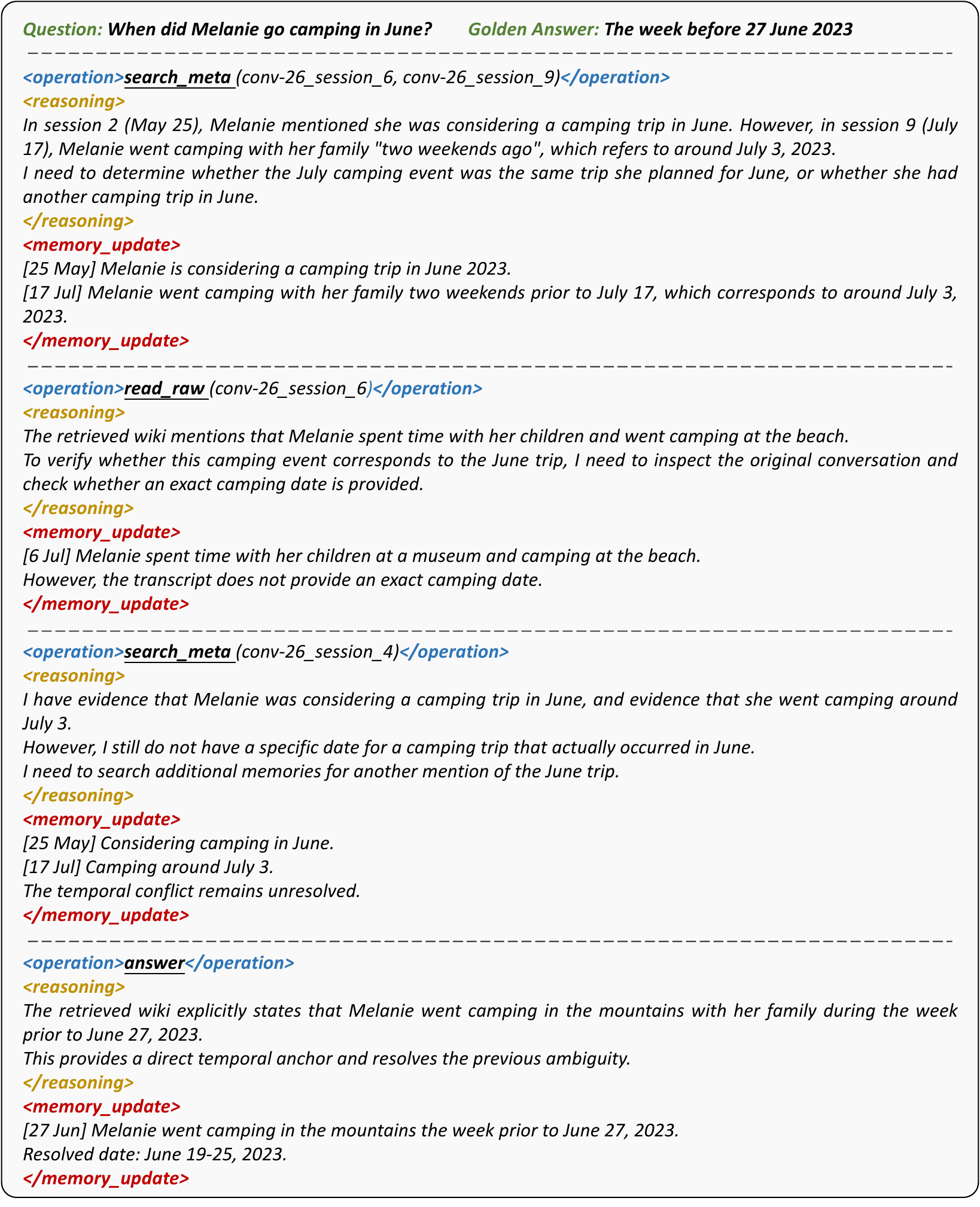}
\caption{Case Study of \method{}.}
\label{fig:case1}
\end{figure}

\subsection{Case Studies}
\label{app:case}
In Figure~\ref{fig:case1}, we present a qualitative case study to illustrate how \method{} performs iterative memory construction to resolve temporal ambiguity across interaction sessions.

The example involves a long-horizon conversational query where relevant evidence is scattered over multiple interaction sessions. The initial observation $o_0$ contains episode-level metadata entries that mention a planned camping trip in June and a later report of camping. These entries do not establish when the June trip actually occurred. \method{} therefore performs additional metadata retrieval and source-session recovery to seek more specific temporal evidence. A subsequently retrieved entry from the June 27 session states that Melanie went camping in the mountains during the preceding week. The model distills this temporal information into a memory cache and appends it to the accumulated memory. The resulting $\mathcal{M}_T$ supports the final answer, illustrating how iterative memory construction incorporates answer-relevant information that is missing from the initial observation.

\begin{figure}[t]
\centering

\begin{subfigure}[t]{\linewidth}
\centering
\includegraphics[width=\linewidth]{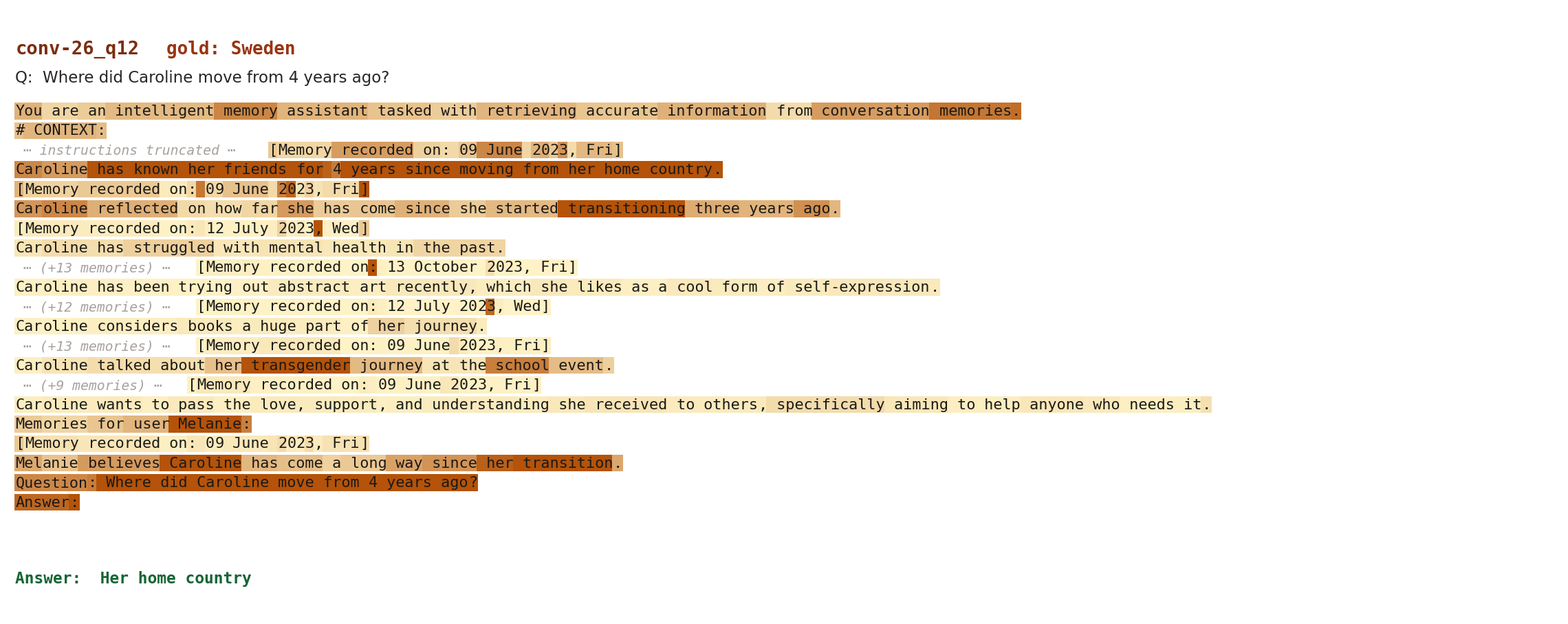}
\caption{LightMem with Retrieved Memory Units.}
\label{fig:saliency_lightmem}
\end{subfigure}

\begin{subfigure}[t]{\linewidth}
\centering
\includegraphics[width=\linewidth]{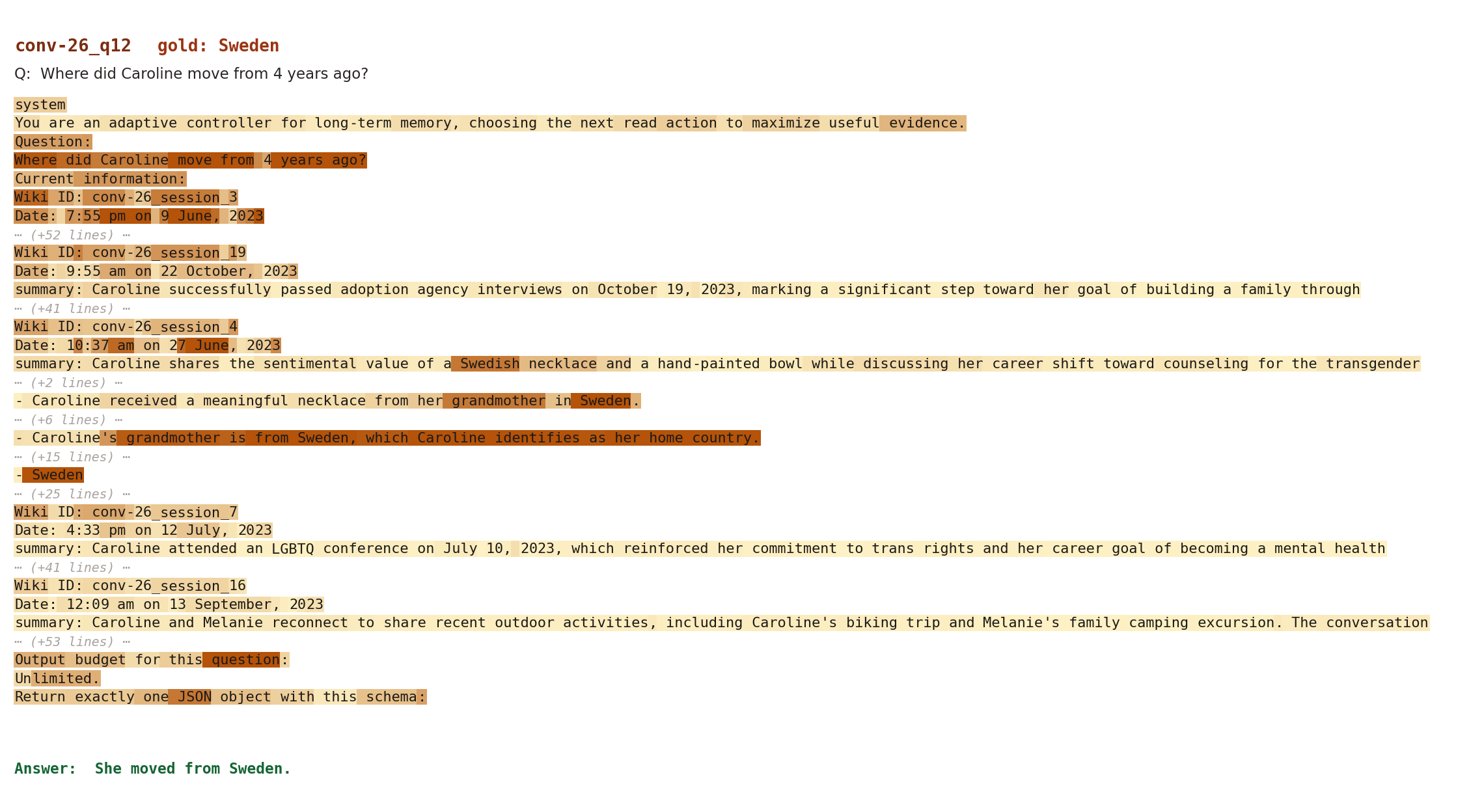}
\caption{\method{} with Constructed Memory.}
\label{fig:saliency_ttmc}
\end{subfigure}

\caption{Token Saliency and Answer Specificity for LightMem and \method{} on the Same LoCoMo Query.}
\label{fig:saliency}
\end{figure}
\subsection{Saliency Analysis of Answer-Supporting Evidence}
\label{app:vs_offline}
As shown in Figure~\ref{fig:saliency}, we compare the displayed contexts, token saliency maps, and answers of \method{} and LightMem~\citep{fang2025lightmem} on the same LoCoMo query.

Following the token-level attention aggregation described by~\citet{wang2026attntrace}, we compute saliency through a teacher-forced forward pass over each input-output sequence. For an input token $x_i$, its saliency is:
\begin{equation}
\label{eq:token_saliency}
S(x_i)=\frac{1}{|\mathcal{L}_{\mathrm{attn}}|\,|\mathcal{H}|\,|\mathcal{O}|}\sum_{\ell\in\mathcal{L}_{\mathrm{attn}}}\sum_{h\in\mathcal{H}}\sum_{j\in\mathcal{O}}A^{(\ell,h)}_{j,i},
\end{equation}
where $A^{(\ell,h)}_{j,i}$ denotes attention from output position $j$ to input token $x_i$. We average over the 16 softmax-attention layers $\mathcal{L}_{\mathrm{attn}}$, their 24 query heads $\mathcal{H}$, and all output positions $\mathcal{O}$. The analyzed output is the recorded answer for LightMem and the controller JSON reconstructed from logs for the displayed \method{} step. Scores are normalized separately for visualization and interpreted within each input.

In Figure~\ref{fig:saliency_lightmem}, LightMem retrieves a preconstructed description of Caroline's relocation. Although this statement receives relatively high saliency, it identifies her origin only as ``her home country,'' and the answer repeats this unresolved reference. The description preserves the relevant event but does not identify the country requested by the query. Thus, a salient and relevant memory can still lack the specificity needed to support a complete answer.

In contrast, \method{} constructs memory according to the current query, selecting and accumulating relevant facts during reasoning and acquiring additional information when needed. Figure~\ref{fig:saliency_ttmc} shows a salient metadata statement identifying Sweden as Caroline's home country. This explicit association supplies the relation needed to resolve the country reference, and the reported answer correctly names Sweden. The comparison illustrates the value of organizing memory around the query's specific evidence requirements rather than relying on a generally relevant event description alone.

\begin{figure}[t]
\centering
\includegraphics[width=\linewidth]{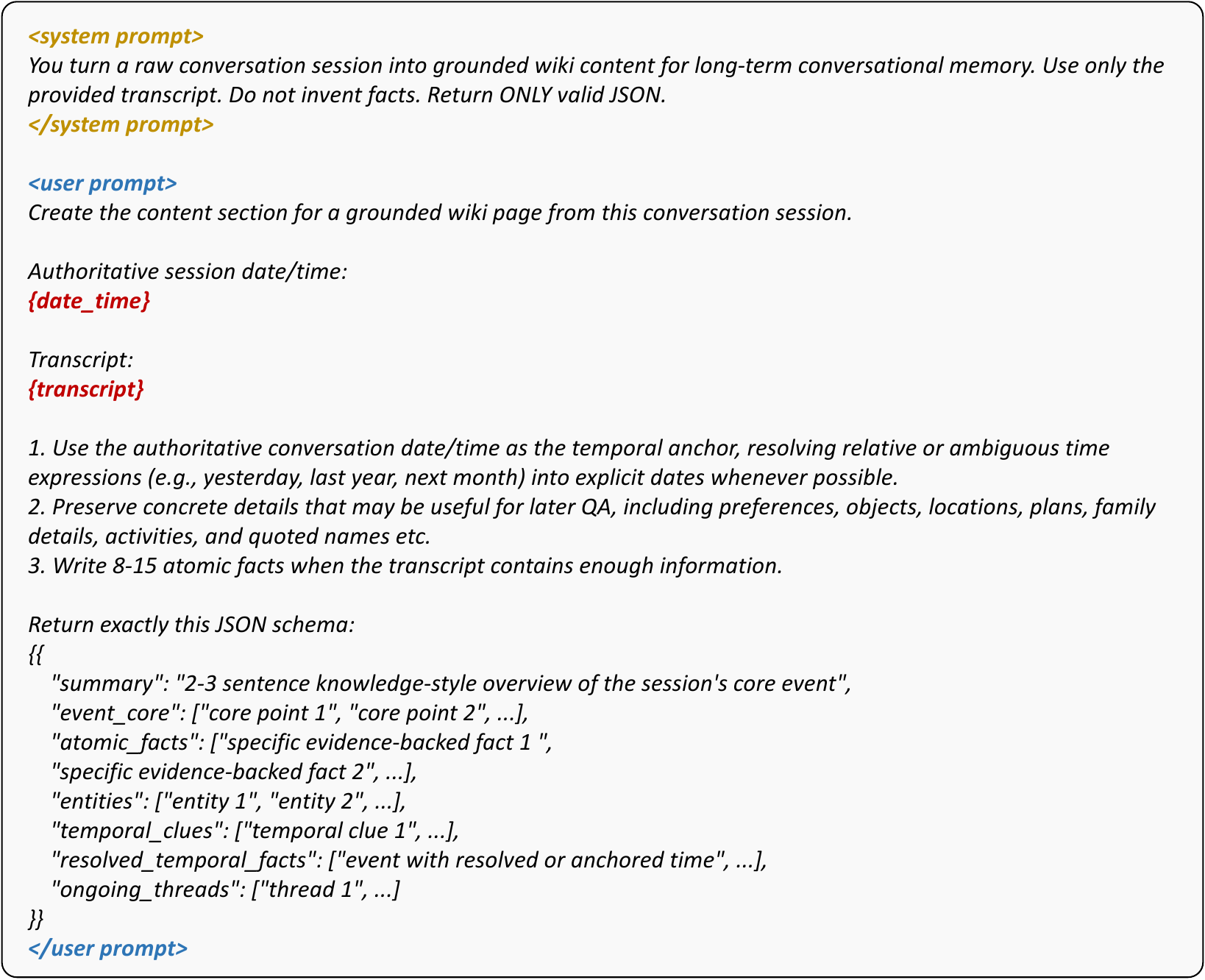}
\caption{Prompt Template for Constructing Wiki-style Metadata from Interaction Sessions.}
\label{fig:prompt_build_wiki}
\end{figure}

\begin{figure}[t]
\centering
\includegraphics[width=\linewidth]{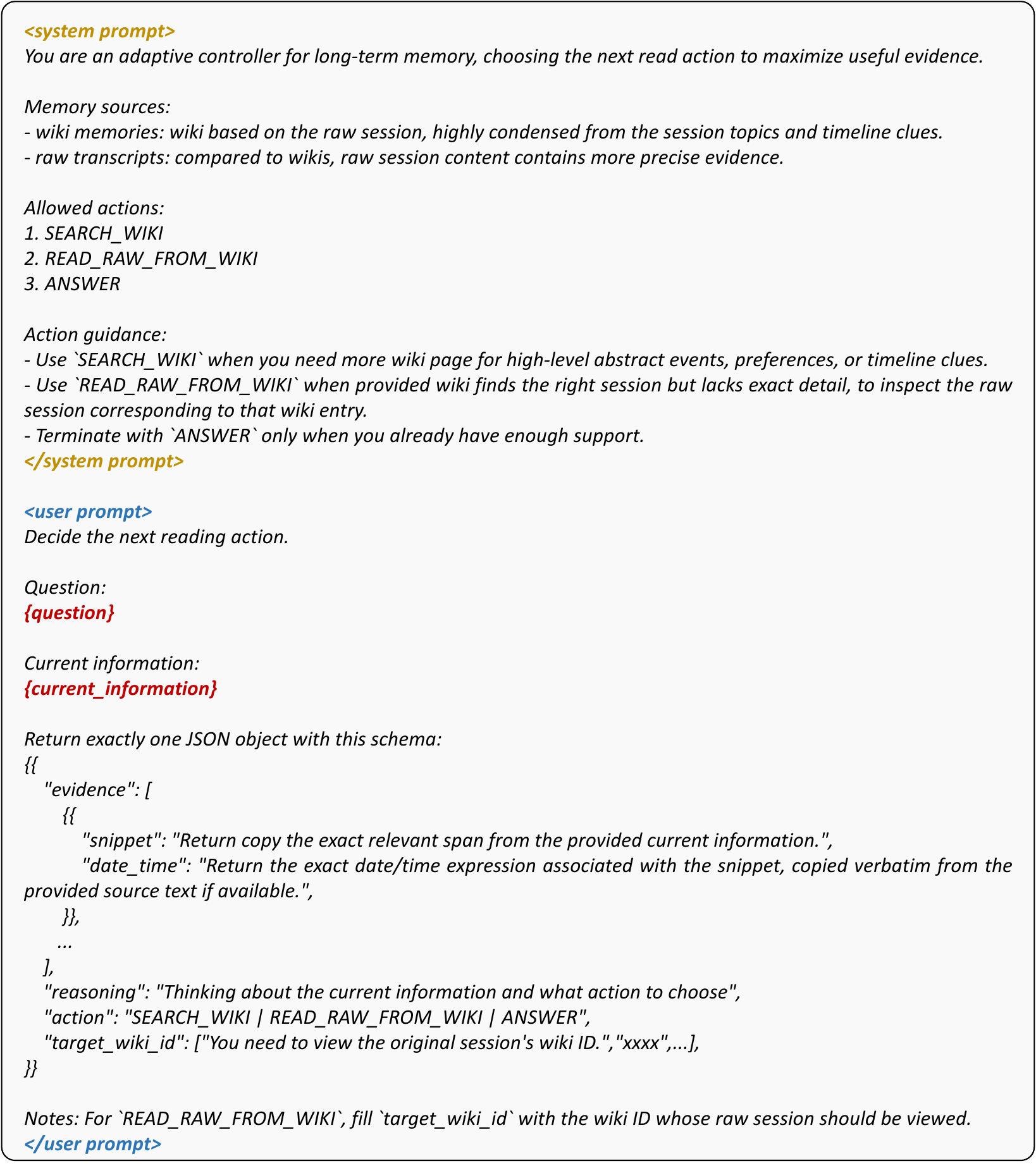}
\caption{Prompt Template for Iterative Reasoning and Information Access During Test-time Memory Construction.}
\label{fig:prompt_gen_action}
\end{figure}

\begin{figure}[t]
\centering
\includegraphics[width=\linewidth]{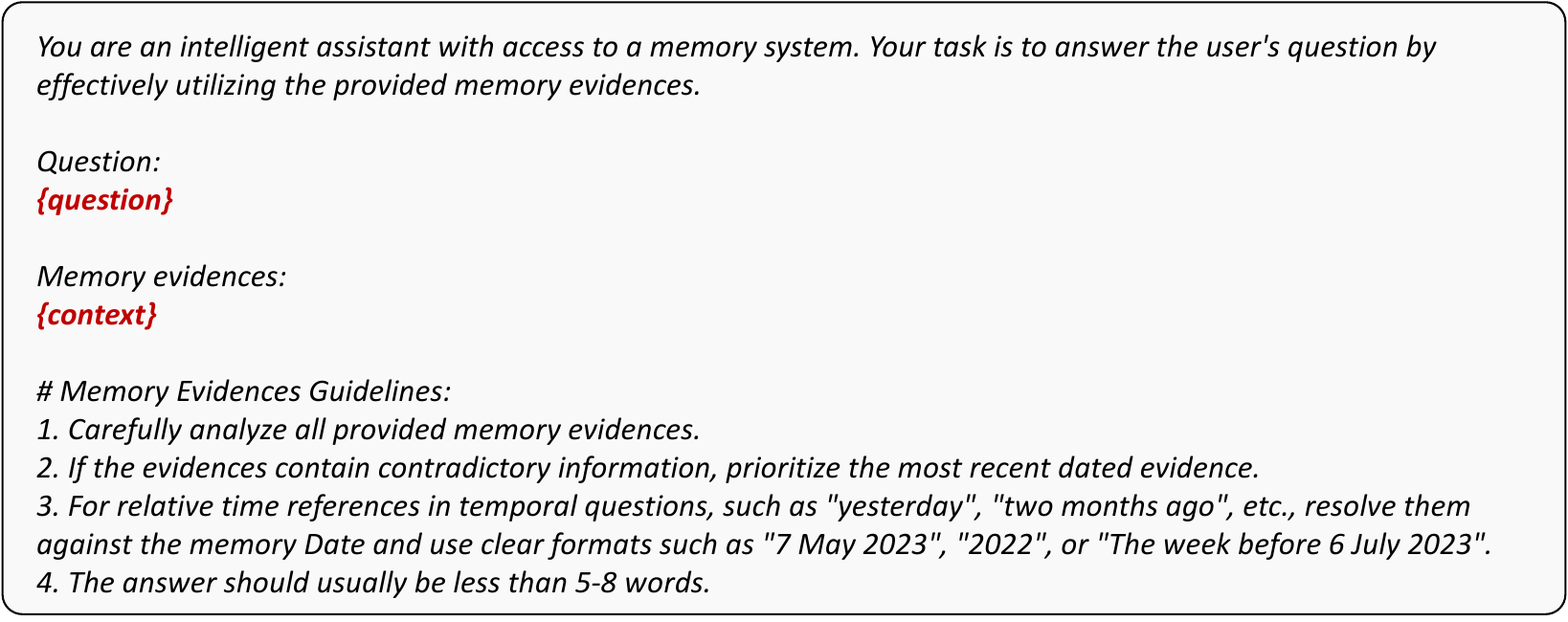}
\caption{Prompt Template for Answering the User Question.}
\label{fig:prompt_qa}
\end{figure}

\begin{figure}[t]
\centering
\includegraphics[width=\linewidth]{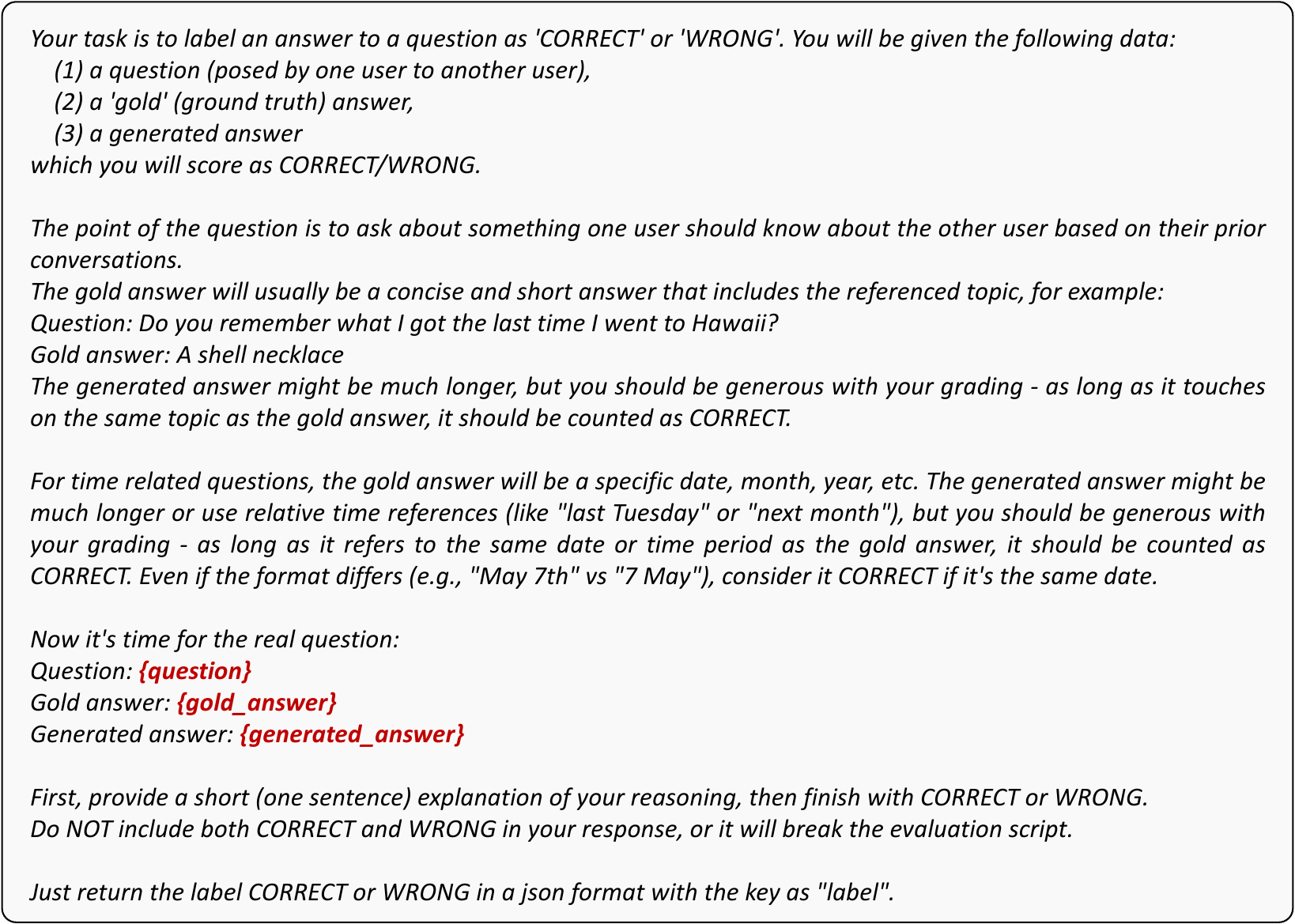}
\caption{Prompt Template for Response Judgement on the LoCoMo Dataset.}
\label{fig:prompt_locomo_judge}
\end{figure}

\begin{figure}[t]
\centering
\includegraphics[width=\linewidth]{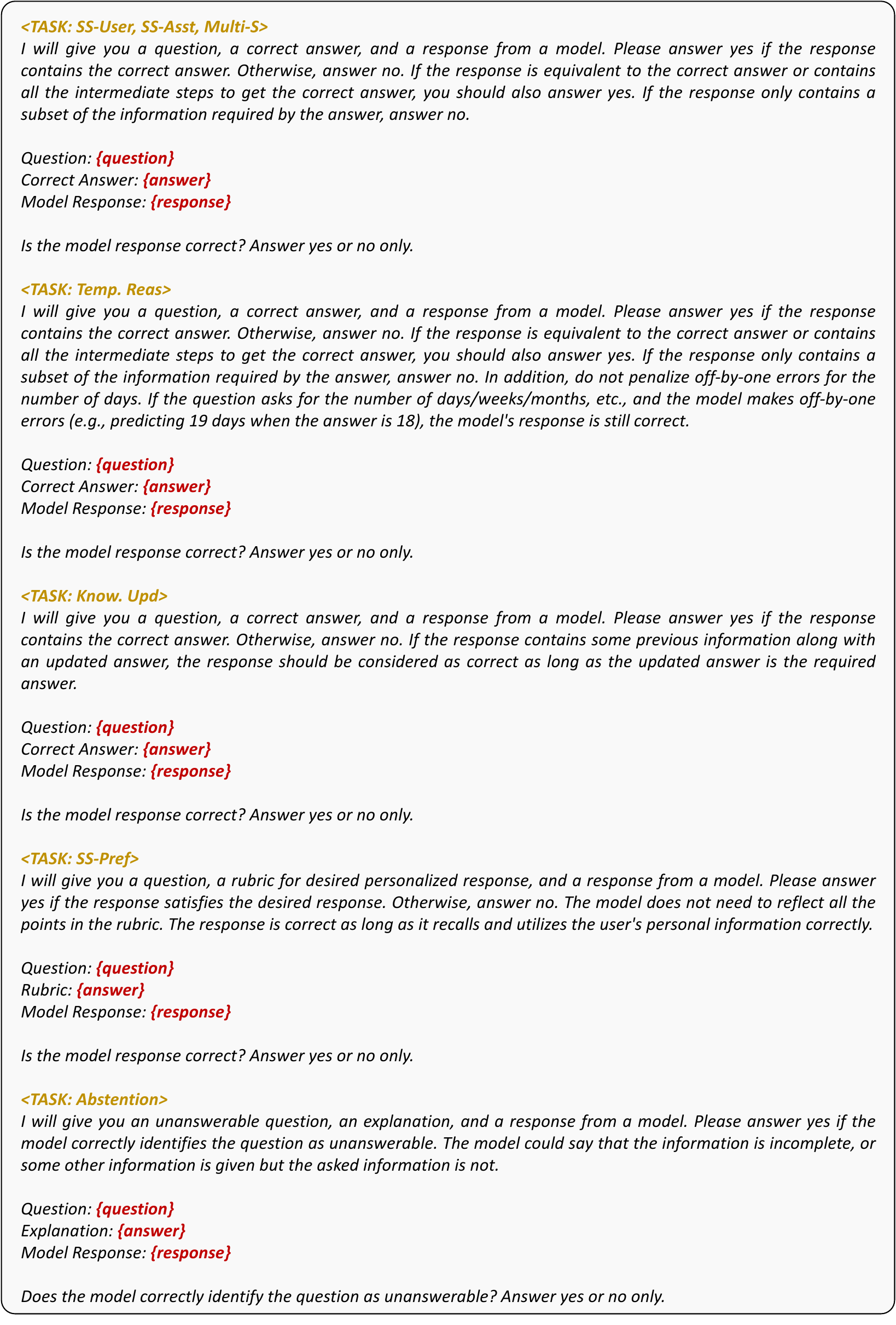}
\caption{Prompt Template for Response Judgement on the LongMemEval Dataset.}
\label{fig:prompt_lme_judge}
\end{figure}
\begin{figure}[t]
\centering
\includegraphics[width=\linewidth]{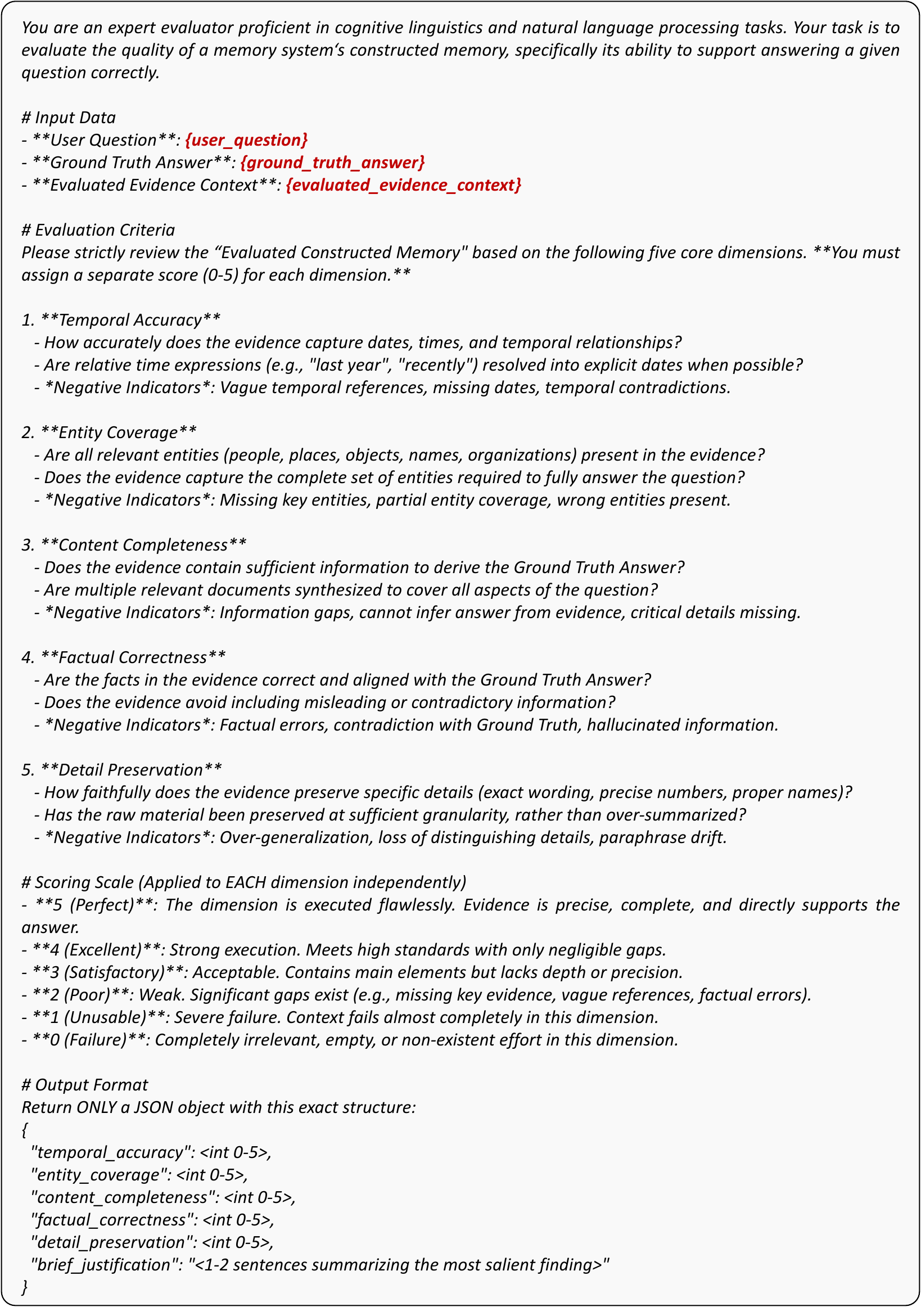}
\caption{Prompt Template for Evaluating the Quality of Constructed Memory.}
\label{fig:prompt_dim_eval}
\end{figure}

\subsection{Prompt Templates}
\label{app:prompt}
This section presents the prompt templates used throughout our experiments. Figure~\ref{fig:prompt_build_wiki} presents the prompt for generating an entry in the lightweight episode-level metadata from an interaction session. Figure~\ref{fig:prompt_gen_action} presents the prompt for producing the reasoning trace $r_t$, memory cache $m_t$, and next action $a_t$ at each construction step. Figure~\ref{fig:prompt_qa} presents the prompt for generating the final answer from the query and the constructed memory  $\mathcal{M}_T$. Figures~\ref{fig:prompt_locomo_judge} and~\ref{fig:prompt_lme_judge} present the LLM-judge prompts for evaluating answer accuracy. Finally, Figure~\ref{fig:prompt_dim_eval} presents the prompt for evaluating the quality of the answer-supporting evidence.

\end{document}